\documentclass{article}
\usepackage{standalone}
\usepackage{microtype}
\usepackage{graphicx}
\usepackage{caption}
\usepackage{booktabs} 
\usepackage{enumitem} 
\usepackage{tikz}
\usepackage{pgfplots} 
\usepgfplotslibrary{fillbetween}
\usetikzlibrary{patterns}
\pgfplotsset{compat=1.17}
\usepackage{tikz-cd}
\usepackage{wrapfig}
\usepackage{tkz-euclide}
\usetikzlibrary{decorations.pathreplacing}
\usepackage{siunitx}
\usepackage{hyperref}

\usepackage{amsmath}
\usepackage{amssymb}
\usepackage{mathtools}
\usepackage{amsthm}

\usepackage[capitalize,noabbrev]{cleveref}

\usepackage[accepted]{icml2025}

\theoremstyle{plain}
\newtheorem{theorem}{Theorem}[section]
\newtheorem{proposition}[theorem]{Proposition}

\newtheorem{corollary}[theorem]{Corollary}
\theoremstyle{definition}
\newtheorem{definition}[theorem]{Definition}

\theoremstyle{remark}

\usepackage[textsize=tiny]{todonotes}

\icmltitlerunning{Multi-Objective Causal Bayesian Optimization}

\begin{document}

\twocolumn[
\icmltitle{Multi-Objective Causal Bayesian Optimization}

\icmlsetsymbol{equal}{*}

\begin{icmlauthorlist}
\icmlauthor{Shriya Bhatija}{tum,cam}
\icmlauthor{Paul-David Zuercher}{cam,alan_turing}
\icmlauthor{Jakob Thumm}{tum}
\icmlauthor{Thomas Bohné}{cam}
\end{icmlauthorlist}

\icmlaffiliation{cam}{Department of Engineering, University of Cambridge, Cambridge, United Kingdom}
\icmlaffiliation{tum}{Department of Computer Engineering, Technical University of Munich, Munich, Germany}
\icmlaffiliation{alan_turing}{The Alan Turing Institute, London, United Kingdom}
\icmlcorrespondingauthor{Shriya Bhatija}{shriya.bhatija@tum.de}
\icmlkeywords{Machine Learning, ICML}
\vskip 0.3in
]

\printAffiliationsAndNotice{}  

\begin{abstract}
In decision-making problems, the outcome of an intervention often depends on the causal relationships between system components and is highly costly to evaluate. In such settings, causal Bayesian optimization (\textsc{cbo}) can exploit the causal relationships between the system variables and sequentially perform interventions to approach the optimum with minimal data. Extending \textsc{cbo} to the multi-outcome setting, we propose \textit{multi-objective causal Bayesian optimization} (\textsc{mo-cbo}), a paradigm for identifying Pareto-optimal interventions within a known multi-target causal graph. We first derive a graphical characterization for potentially optimal sets of variables to intervene upon. Showing that any \textsc{mo-cbo} problem can be decomposed into several traditional multi-objective optimization tasks, we then introduce an algorithm that sequentially balances exploration across these tasks using relative hypervolume improvement.  The proposed method will be validated on both synthetic and real-world causal graphs, demonstrating its superiority over traditional (non-causal) multi-objective Bayesian optimization in settings where causal information is available.

\end{abstract}
\section{Introduction}\label{chap:intro}

\begin{figure}
    \centering
    \vspace{-1.2cm}
    \includegraphics{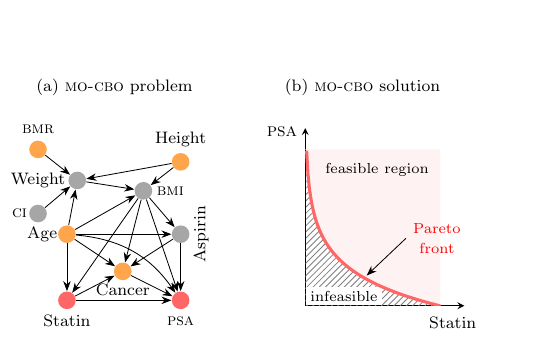}
    \vspace{-0.9cm}
    \caption{\textsc{mo-cbo} problem in healthcare. (a) Causal graph where red, grey, and orange nodes depict target, manipulative, and non-manipulative variables, respectively. (b) The solution consists of interventions that yield optimal trade-offs between the targets.}
    \label{fig:health_dag}
\end{figure}

Decision-making problems arise in a variety of domains, such as healthcare, manufacturing or public policy, and involve manipulating variables to obtain an outcome of interest. In many such domains, interventions are inherently costly, and practical applications are subject to budgetary constraints. Moreover, these systems are often governed by causal mechanisms, which can be exploited to efficiently approach optimal outcomes in a targeted and cost-efficient manner. A well-established strategy for optimizing such expensive-to-evaluate black-box functions is Bayesian optimization \cite{BO_review}, but it cannot leverage the causal structure between its input variables. To this end, causal Bayesian optimization (\textsc{cbo}) \cite{CBO} was introduced to generalize Bayesian optimization to settings where causal information is available. 
While existing \textsc{cbo} variants focus on optimizing a single objective, real-world systems often require simultaneous optimization of multiple outcomes. Here, the aim is to establish optimal trade-offs between objectives, a so called Pareto front, instead of identifying a single objective's global optimum. As an example, consider the graph in \cref{fig:health_dag} (a), depicting the causal relationships between prostate specific antigen (\textsc{psa}) and its risk factors \cite{ferro_healthcare}. For patients sensitive to Statin medications, the aim is to determine how to reduce the relevant risk factors to minimize both Statin and \textsc{psa} simultaneously. \autoref{fig:health_dag} (b) shows how the optimal trade-offs between the targets could behave.

\begin{figure*}[t]
    \begin{tikzpicture}[scale=0.45,
    ->, 
     >=Stealth, 
    every edge/.style={draw}, 
    box/.style={draw, fill=white,
                minimum width=2.4cm,
                minimum height=1.5cm
                },
    ellipsis/.style={font=\large}
    ovbx/.pic={\foreach \c in {0,.1,.2}
                    \node[box] (X\c) at (\c,-\c) {};
                \coordinate (-WE) at (X0.west);
              },
    ]

    \node[font=\footnotesize] at (5,-0.8) {\textsc{mo-cbo} \textbf{problem}};
    \node[font=\footnotesize] at (14.5,-0.8) {\textbf{Search space reduction}};
    \node[font=\footnotesize] at (14.5,-1.6) {[Sec. \ref{subsec:search_space_reduction}]};
    \node[font=\footnotesize] at (24.1,-0.8) {\textbf{Solve the local problems}};
    \node[font=\footnotesize] at (24.1,-1.6) {[Sec. \ref{subsec:mo_cbo_algorithm}]};
    \node[font=\footnotesize] at (33.2,-0.8) {\textbf{Build Pareto front}};

    \draw[double,thick] (10.0,-7) -- (10.9,-7); 
    \draw[double,thick] (18.0,-7) -- (19.0,-7); 
    \draw[double,thick] (28.9,-7) -- (29.8,-7); 

    \node[draw, minimum width=4.5cm, minimum height=1.5cm, outer sep=0, anchor=north west] (main_problems_box) at (0, -1.8) {};
    
    \begin{scope}[shift={(main_problems_box.south)}, xshift=1.0cm, yshift=6.8cm]
        \node[circle, inner sep=2.5pt, fill=gray!70] (X4) at (-3.6, -4.5) {};
        \node[font=\scriptsize] (X4_ann) at (-4.2, -5.0) {$X_4$};
        \node[circle, inner sep=2.5pt, fill=gray!70] (X1) at (-1.8, -4.5) {};
        \node[font=\scriptsize] (X1_ann) at (-2.4, -5.0) {$X_1$};
        \node[circle, inner sep=2.5pt, fill=gray!70] (X2) at (-1.8, -5.6) {};
        \node[font=\scriptsize] (X2_ann) at (-1.8, -6.3) {$X_2$};
        \node[circle, inner sep=2.5pt,fill=red!60] (Y1) at (0.0, -4.5) {};
        \node[font=\scriptsize] (Y1_ann) at (0.7, -4.5) {$Y_1$};
        \node[circle, inner sep=2.5pt, fill=red!60] (Y2) at (0.0, -5.6) {};
        \node[font=\scriptsize] (Y2_ann) at (0.0, -6.3) {$Y_2$};
        \node[circle, inner sep=2.5pt,  fill=gray!70] (X3) at (1.8, -5.6) {};
        \node[font=\scriptsize] (X3_ann) at (1.8, -6.3) {$X_3$};
        \draw (X1) edge (Y1);
        \draw (X2) edge (Y1);
        \draw (X2) edge (Y2);
        \draw (X3) edge (Y2);
        \draw (X4) edge (X1);
        \draw[dashed, <->, bend left=35] (X4) to (Y1);
    \end{scope}

    \node[draw, minimum width=4.5cm, minimum height=2.7cm, outer sep=0, anchor=north west] (local_problems_box) at (0, -5.4) {};

    \node[align=center, font=\small, anchor=north, text width=4.5cm] at (local_problems_box.north) {Full search space has $|\mathbb{P}(\mathbf{X})| = 2^{|\mathbf{X}|}$ local problems};

    \begin{scope}[shift={(local_problems_box.south)}, xshift=1.0cm, yshift=2.4cm] 
        \node[draw, fill=white, minimum width=1.5cm, minimum height=0.9cm] at (0,0) {};
        \node[draw, fill=white, minimum width=1.5cm, minimum height=0.9cm] at (-0.2,-0.2) {};
        \node[draw, fill=white, minimum width=1.5cm, minimum height=0.9cm] at (-0.4,-0.4) {};
        \node[draw, fill=white, minimum width=1.5cm, minimum height=0.9cm] at (-0.6,-0.6) {};
        \node[draw, fill=white, minimum width=1.5cm, minimum height=0.9cm] at (-0.8,-0.8) {};
        \node[draw, fill=white, minimum width=1.5cm, minimum height=0.9cm] at (-1.0,-1.0) {};
        \node[draw, fill=white, minimum width=1.5cm, minimum height=0.9cm] at (-1.2,-1.2) {};

        \node[font=\scriptsize] at (-1.2,-0.8) {local};
        \node[font=\scriptsize] at (-1.2,-1.5) {problem 1};
        \draw[-,decorate, decoration={brace, amplitude=7pt}, rotate=135] (1.8,2.9) -- (2.1,-0.0); 
        \node[font=\footnotesize] at (-3.5,0.9) {16};
    \end{scope}

    \node[box, minimum width=3.2cm, minimum height=4cm] at (14.5, -7) {};
    \node[box, minimum width=2.7cm, minimum height=1.8cm] at (14.5, -4.8) {};
    \node[font=\footnotesize] at (14.5,-3.3) {local problem 1};
    \node[circle, inner sep=2.5pt, fill=gray!70] (X1) at (12.9, -4.5) {};
    \node[font=\scriptsize] (X1_ann) at (12.2, -4.5) {$X_1$};
    \node[circle, inner sep=2.5pt, fill=gray!70] (X2) at (12.9, -5.6) {};
    \node[font=\scriptsize] (X2_ann) at (12.9, -6.3) {$X_2$};
    \node[circle, inner sep=2.5pt,fill=red!60] (Y1) at (14.7, -4.5) {};
    \node[font=\scriptsize] (Y1_ann) at (15.4, -4.5) {$Y_1$};
    \node[circle, inner sep=2.5pt, fill=red!60] (Y2) at (14.7, -5.6) {};
    \node[font=\scriptsize] (Y2_ann) at (14.7, -6.3) {$Y_2$};
    \node[circle, inner sep=2.5pt,  fill=gray!70] (X3) at (16.5, -5.6) {};
    \node[font=\scriptsize] (X3_ann) at (16.5, -6.3) {$X_3$};
    \draw (X1) edge (Y1);
    \draw (X2) edge (Y1);
    \draw (X2) edge (Y2);
    \draw (X3) edge (Y2);

    \node[box, minimum width=2.7cm, minimum height=1.8cm] at (14.5, -9.1) {};
    \node[font=\footnotesize] at (14.5,-7.6) {local problem 2};
    \node[circle, inner sep=2.5pt, fill=gray!70] (X2) at (12.9, -9.9) {};
    \node[font=\scriptsize] (X2_ann) at (12.9, -10.6) {$X_2$};
    \node[circle, inner sep=2.5pt,fill=red!60] (Y1) at (14.7, -8.8) {};
    \node[font=\scriptsize] (Y1_ann) at (15.4, -8.8) {$Y_1$};
    \node[circle, inner sep=2.5pt, fill=red!60] (Y2) at (14.7, -9.9) {};
    \node[font=\scriptsize] (Y2_ann) at (14.7, -10.6) {$Y_2$};
    \node[circle, inner sep=2.5pt,  fill=gray!70] (X3) at (16.5, -9.9) {};
    \node[font=\scriptsize] (X3_ann) at (16.5, -10.6) {$X_3$};
    \draw (X2) edge (Y1);
    \draw (X2) edge (Y2);
    \draw (X3) edge (Y2);

    \node[box, minimum width=4.5cm, minimum height=4cm] at (23.9, -7) {};
    \node[font=\footnotesize] at (23.9, -3.2) {\textsc{Causal ParetoSelect}};
    \node[box, minimum width=1.5cm, minimum height=0.9cm] at (24.8, -6.8) {};
    \node[box, minimum width=1.5cm, minimum height=0.9cm] at (24.2, -7.4) {};
    \node[font=\scriptsize] at (24.2,-7) {local};
    \node[font=\scriptsize] at (24.2,-7.7) {problem 1};
    \node[font=\scriptsize] at (20.3, -7.2) {1. Select};
    \node[font=\scriptsize] at (26.4, -9.5) {2. Optimize};
    \node[font=\scriptsize] at (26.7, -10.1) {(\textsc{dgemo})};
    \draw[->] (21.6,-7.2) -- (22.5,-7.2); 
    \draw[->] (25.3,-8.6) arc[start angle=180, end angle=430, radius=0.5cm]; 
    \draw[->] (24.2,-4) -- (24.2,-5.6);   
    \draw[->] (28.5,-4.6) -- (24.2,-4.6); 
    \draw[-] (28.5,-4.6) -- (28.5,-10.5); 
    \draw[->] (24.2,-8.4) -- (24.2,-11.3);
    \draw[-] (28.5,-10.5) -- (24.2,-10.5); 
    \node[circle,fill=black,inner sep=1] at (24.2,-10.5) {};

    \node[box, minimum width=3.1cm, minimum height=4cm] at (33.2, -7) {};
    \draw[-,ultra thick, red!60] (30.9, -9.5) -- (32.2, -9.5);
    \node[font=\scriptsize] at (31.6, -9.9) {causal};
    \node[font=\scriptsize] at (31.6, -10.5) {Pareto};
    \node[font=\scriptsize] at (33.0, -10.5) {front};
    \draw[-,ultra thick, gray!70] (32.6, -3.6) -- (33.8, -3.6);
    \node[font=\scriptsize] at (34.3, -4.0) {Pareto fronts of};
    \node[font=\scriptsize] at (34.25, -4.7) {local problems};

    \begin{axis}[
        scale only axis,
        no markers,
        domain=0:5,
        samples=100,
        axis lines=center,
        axis line style={line width=1.5pt},
        ticks=none,
        xmin=0, xmax=5.3,            
        ymin=, ymax=10,   
        xlabel={$Y_1$},
        ylabel={$Y_2$},
        xlabel style={yshift=5pt, xshift=15pt, font=\fontsize{15}{17}\selectfont},
        ylabel style={yshift=10pt, xshift=-23pt, font=\fontsize{15}{17}\selectfont},
        height=7.7cm,                  
        width=5.3cm,
        at={(3050,-110)}]
        \addplot[red!60, -, line width=4pt, restrict x to domain=0.2:2.35] {5.8 - 0.95*ln(x)}; 
        \addplot[red!60, -, line width=4pt, restrict x to domain=2.25:4.5] {8.3 - 4*ln(x)}; 
        \addplot[gray!80, -, thick, line width=3pt, restrict x to domain=0.2:4.5] {8.5 - 4*ln(x)};  
        \addplot[gray!80, -, thick, line width=3pt, restrict x to domain=0.2:4.5] {6 - ln(x)};
    \end{axis}
    \vspace{-0.2cm}
\end{tikzpicture}
    \caption{Overview of the \textsc{mo-cbo} methodology.}
    \label{fig:methodology_overview}
\end{figure*}
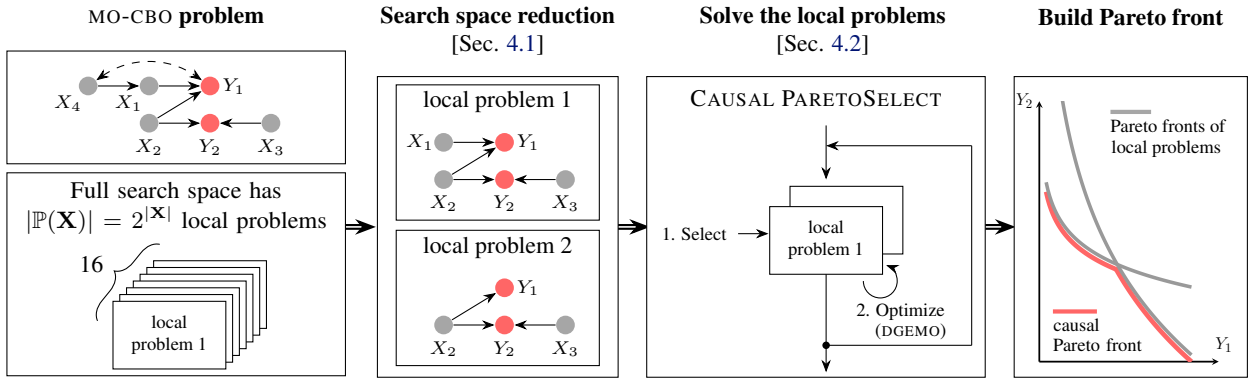

We propose \textit{multi-objective causal Bayesian optimization} (\textsc{mo-cbo}) as a method that generalizes \textsc{cbo} to such optimization problems involving multiple outcome variables. A visualization of our methodology is given in \cref{fig:methodology_overview}, highlighting some of the key contributions of our work:

\begin{enumerate}[left=0pt]
    \item We formally define \textsc{mo-cbo} as a new class of optimization problems.
    \item We provide a formal method to reduce the search space of \textsc{mo-cbo} problems, thereby disregarding sets of variables that are not worth intervening upon based on the graph topology. 
    \item We propose \textsc{Causal ParetoSelect}, an efficient algorithm for the exploration of multiple intervention strategies in parallel, guided by a custom acquisition function.
    \item We experimentally show that our approach can surpass traditional (non-causal) multi-objective Bayesian optimization in settings where causal relationships are known, achieving more cost-effective, diverse, and accurate solutions.
\end{enumerate}

To the best of our knowledge, there exists no other multi-objective optimization method in the literature that can consider the causal structure. We prove that our \textsc{mo-cbo} methodology establishes optimal trade-offs between target variables,  which state-of-the-art methods cannot reach for all experimental setups.

\subsection{Related Work}
We combine multi-objective Bayesian optimization (\textsc{mobo}) and techniques from causal inference to achieve multi-objective causal Bayesian optimization.
Developments in \textsc{mobo} range from single-point methods \cite{ParEGO, PAL, USeMO} and batch methods \cite{MOEA/D-EGO, dgemo, qNEHVI} to the derivations of efficient acquisition functions \cite{EHVI,PES}. 
Most relevant for our work is \textsc{dgemo} \cite{dgemo}, a well-established \textsc{mobo} algorithm that employs a novel batch selection strategy emphasizing on sample diversity in the input space.

Leveraging tools from causal inference to make causally-informed decisions is a field called causal decision-making. Within this field, there is a growing body of research specifically focused on advancing \textsc{cbo} \cite{CBO}. These advancements include extensions such as constrained \textsc{cbo} \cite{pmlr-v202-aglietti23a}, dynamic \textsc{cbo} \cite{NEURIPS2021_577bcc91}, and various other variants \cite{pmlr-v206-branchini23a, pmlr-v216-gultchin23a, sussex2023modelbasedcausalbayesianoptimization, sussex2023adversarialcausalbayesianoptimization, cbo_exogenous_learning, causal_elicitation}. They are all designed to optimize a single target variable, rendering them infeasible for applications with multiple outcomes.

\section{Preliminaries}

In this paper, random variables and their realizations are denoted in the upper and lower case, respectively. Sets and vectors are written in bold. For a set $\mathbf{X}$, its power set is denoted as $\mathbb{P}(\mathbf{X})$.

\subsection{Multi-objective Bayesian optimization}\label{subsec:prelim_mobo}
Multi-objective Bayesian optimization simultaneously minimizes (or, maximizes) a set of black-box objectives $f_1, \dots, f_m: \mathbf{X} \rightarrow \mathbb{R}$ with minimal function evaluations. Due to potential conflicts between objectives, it aims to find trade-off solutions, known as Pareto optima:

\begin{definition}[Pareto optimality]
    A point $x \in \mathbf{X}$ is called \textit{Pareto-optimal} if there is no other $x' \in \mathbf{X}$ such that $f_i(x) \geq f_i(x')$ for all $1\leq i \leq m$, and $f_i(x) > f_i(x')$ for at least one $1\leq i \leq m$. The set of Pareto-optimal points in $\mathbf{X}$ is called \textit{Pareto set}, denoted $\mathcal{P}_s$. The \textit{Pareto front} is the image of the Pareto set under the objective functions, given by $\mathcal{P}_f = \{ \mathbf{f}(x) = (f_1(x),\dots,f_m(x)) \ | \ x \in \mathcal{P}_s \}$.
\end{definition}

At each iteration of a \textsc{mobo} algorithm, prior data is used to fit a \textit{surrogate model} of the objectives, for which Gaussian processes \cite{gps} are predominantly used. Based on the surrogates, an approximation $\tilde{\mathcal{P}}_f$ of the Pareto front is computed. The next point/batch to be evaluated is determined by maximizing the \textit{acquisition function}, which estimates the utility of evaluating the objectives at a given point/batch. The most commonly used acquisition function is the hypervolume indicator $\mathcal{H}$ \citep{dgemo_hypervolume}. The larger the hypervolume, the better $\tilde{\mathcal{P}}_f$  approximates the true Pareto front. To determine how much the hypervolume would increase if a batch of samples $\mathbf{B} \subseteq \mathbf{X}$ was added to the current approximation, hypervolume improvement is used:
\begin{equation}
    \text{HVI}(\mathbf{f}(\mathbf{B}), \tilde{\mathcal{P}}_f)= \mathcal{H} (\tilde{\mathcal{P}}_f \cup \mathbf{f}(\mathbf{B})) - \mathcal{H}(\tilde{\mathcal{P}}_f).
\end{equation}

Since \textsc{dgemo} \citep{dgemo} is the relevant \textsc{mobo} algorithm for our work, we briefly describe its batch selection strategy. It considers hypervolume improvement as well as sample diversity in the input space. To this end, the so called diversity regions $\mathcal{R}_1,\dots,\mathcal{R}_K$ are constructed by using the current Pareto front approximation to group the optimal points based on their performance properties in the input space. Formally, a batch is chosen as follows:
\begin{align}\label{eq:mo_cbo.dgemo_batch_selection}
    \mathbf{B} = &\underset{\mathbf{B} \subseteq \mathbf{X}, |\mathbf{B}| = B}{\text{arg max }} \text{HVI}(\mathbf{f}(\mathbf{B}), \tilde{\mathcal{P}}_f) \notag \\
    & \hspace{0.7cm} \text{ s.t. } \underset{1 \leq k \leq K}{\text{max}} \delta_k(\mathbf{B}) - \underset{1 \leq k \leq K}{\text{min}} \delta_k(\mathbf{B}) \leq 1,
\end{align}
where $B$ denotes the batch size and the functions $\delta_{k}(\cdot)$ are defined as the number of elements from $\mathbf{B}$ that belong to $\mathcal{R}_k$. For the complete selection algorithm, we refer to \citet{dgemo}.

\subsection{Causality}
\paragraph{Graph notation} A graph $\mathcal{G} = (\mathbf{V}, \mathbf{E})$ is defined by a finite vertex set $\mathbf{V}$ and an edge set $\mathbf{E} \subseteq \mathbf{V} \times \mathbf{V}$, containing ordered pairs of distinct vertices. The subgraph of $\mathcal{G}$ restricted to $\mathbf{V}' \subseteq \mathbf{V}$ is given by $\mathcal{G}[\mathbf{V}'] = ( \mathbf{V}', \mathbf{E}[\mathbf{V}'])$, where $\mathbf{E}[\mathbf{V}'] = \{(i,j) \in \mathbf{E} \ | \ i,j \in \mathbf{V}'\}$. 
For $V \in \mathbf{V}$, the set of its parents, ancestors and descendants in $\mathcal{G}$ is denoted as $\text{pa}(V)_{\mathcal{G}}$, $\text{an}(V)_{\mathcal{G}}$, and $\text{de}(V)_{\mathcal{G}}$, respectively. 
Here, no vertex is a parent, an ancestor, or a descendant of itself. 
Conversely, with a capital letter, this notation is extended to include the argument in the result, i.e., $\text{Pa}(V)_{\mathcal{G}} = \text{pa}(V)_{\mathcal{G}} \cup \{ V \}$. 
Moreover, we define these relations for sets of variables $\mathbf{V}' \subseteq \mathbf{V}$, i.e., $\text{pa}(\mathbf{V}')_{\mathcal{G}} = \bigcup_{V \in \mathbf{V'}} \text{pa}(V)_{\mathcal{G}}$ and $\text{Pa}(\mathbf{V}')_{\mathcal{G}} = \bigcup_{V \in \mathbf{V'}} \text{Pa}(V)_{\mathcal{G}}$. 
Equivalent conventions apply to the ancestor and descendant relationships.

\paragraph{Structural causal models} Let $\langle \mathbf{V}, \mathbf{U}, \mathbf{F}, P(\textbf{U}) \rangle$ be a structural causal model (\textsc{scm}) \cite{Pearl_00} and $\mathcal{G}$ its associated acyclic graph that encodes the underlying causal mechanisms. 
Specifically, $\mathbf{U}$ is a set of independent exogenous random variables distributed according to the probability distribution $P(\mathbf{U})$, $\mathbf{V}$ is a set of endogenous random variables, and $\mathbf{F} = \{f_V\}_{V \in \mathbf{V}}$ is a set of deterministic functions such that $V = f_V(\text{pa}(V)_{\mathcal{G}}, \mathbf{U}^V)$, where $\mathbf{U}^{V} \subseteq \mathbf{U}$ is the set of exogenous variables affecting $V \in \mathbf{V}$. The set $\mathbf{U}^V \cap \mathbf{U}^W$ consists of unobserved confounders between $V,W \in \mathbf{V}$, which are the exogenous variables influencing both $V$ and $W$. Within $\mathbf{V}$ there are three different types of variables to be distinguished: non-manipulative variables $\mathbf{C}$ that cannot be modified, treatment variables $\mathbf{X}$ which can be set to specific values, and output variables $\mathbf{Y} = \{Y_1,\dots,Y_m\}$ which represent the outcome of interest. We consider only real-valued \textsc{scm}s, where all endogenous variables have continuous domains. 
For $\mathbf{X}_s \subseteq \mathbf{X}$, $\text{CC}(\mathbf{X}_s)_{\mathcal{G}}$ refers to the c-component of $\mathcal{G}$ \citep{do_calculus.proof_3}, which, in this context, is the maximal set of variables that includes $\mathbf{X}_s$ and is connected via unobserved confounders. The joint distribution of $\mathbf{V}$, which is determined by $P(\mathbf{U})$, is referred as \textit{observational distribution} and denoted $P(\mathbf{V})$. 

\paragraph{Interventions} A set $\mathbf{X}_s \in \mathbb{P}(\mathbf{X})$ is also called an intervention set. The interventional domain of an intervention set is given as $\mathcal{D}(\mathbf{X}_s)= \times_{X \in \mathbf{X}_s} \mathcal{D}(X)$ and determines the feasible values of $\mathbf{X}_s$. An \textit{intervention} on $\mathbf{X}_s$ involves replacing the structural equations $f_{X}$ with a constant $x$, for all $X \in \mathbf{X}_s$. This action is denoted with the do-operator $\text{do}(\mathbf{X}_s = \mathbf{x}_s)$, where the \textit{intervention value} is $\mathbf{x}_s \in \mathcal{D}(\mathbf{X}_s)$. 
The graph $\mathcal{G}_{\overline{\mathbf{X}}_s}$ represents this intervention and is obtained by removing the incoming edges into $\mathbf{X}_s$. 
The observational distribution of $\mathcal{G}_{\overline{\mathbf{X}}_s}$ is denoted as $P(\mathbf{V} | \text{do}(\mathbf{X}_s=\mathbf{x}_s))$
and called \textit{interventional distribution}. 
For $\mathbf{X}_s =\varnothing$, no intervention is performed and the observational and interventional distributions coincide. The tuple $(\mathbf{X}_s,\mathbf{x}_s)$ is referred to as an \textit{intervention set-value} pair. Given two sets $\mathbf{X}_s, \mathbf{X}_s' \subseteq \mathbf{X}$ and $\mathbf{x}_s \in \mathcal{D}(\mathbf{X}_s)$, we write by $\mathbf{x}_s[\mathbf{X}_s']$ the values of $\mathbf{x}_s$ corresponding to $\mathbf{X}_s \cap \mathbf{X}_s'$.

\section{The \textsc{mo-cbo} 
Problem}\label{sec:mo_cbo}

In our setting, we require that the causal relationships encoded in $\mathcal{G}$ are known while the underlying parametrizations, i.e., $\mathbf{F}$ and $P(\mathbf{U})$, can be unknown. This restricted information is denoted as $\langle \mathcal{G},\mathbf{Y}, \mathbf{X}, \mathbf{C}\rangle$ and the assumption allows generalization across systems that share the same causal structure. 

A \textsc{mo-cbo} problem aims to identify intervention set-value pairs $(\mathbf{X}_s, 
\mathbf{x}_s)$ that can minimize all target variables in $\mathbf{Y}$ simultaneously. The outcomes of an intervention $\text{do}(\mathbf{X}_s=\mathbf{x}_s)$ are captured as the expected values:
\begin{equation}
\label{eq:exp_target}
    \mathbb{E}_{P(Y_i | \text{do}(\mathbf{X}_s=\mathbf{x}_s))}[Y_i] := \mu_i(\mathbf{X}_s,\mathbf{x}_s),
\end{equation}
where $P(Y_i | \text{do}(\mathbf{X}_s=\mathbf{x}_s))$ denotes the interventional distribution of $Y_i$, for all $i=1,\dots,m$.  
We write $\boldsymbol{\mu}(\mathbf{X}_s,\mathbf{x}_s) = (\mu_1(\mathbf{X}_s,\mathbf{x}_s),\dots,\mu_m(\mathbf{X}_s,\mathbf{x}_s))$ for the vector notation. Since opposing causal relationships among the variables can cause conflicting single-objective optima, we consider multi-objective optimization settings, where the aim is to find Pareto-optimal solutions to establish trade-offs between the objective functions. This motivates the application of Pareto optimality to intervention set-value pairs:

\begin{definition}[Pareto-optimal intervention set-value pair]
    Given $\mathcal{S} \subseteq \mathbb{P}(\mathbf{X})$, an intervention set-value pair $(\mathbf{X}_s, \mathbf{x}_s)$ with $\mathbf{X}_s \in \mathcal{S}$, $\mathbf{x}_s \in \mathcal{D}(\mathbf{X}_s)$ is called \textit{Pareto-optimal for $\mathcal{S}$}, if there is no other intervention set-value pair $(\mathbf{X}_s', \mathbf{x}_s')$ with $\mathbf{X}_s' \in \mathcal{S}$, $\mathbf{x}_s' \in \mathcal{D}(\mathbf{X}_s')$ such that $\mu_i(\mathbf{X}_s', \mathbf{x}_s') \leq \mu_i(\mathbf{X}_s, \mathbf{x}_s)$ for all $1 \leq i \leq m$, and $\mu_i(\mathbf{X}_s', \mathbf{x}_s') < \mu_i(\mathbf{X}_s, \mathbf{x}_s)$ for at least one $1 \leq i \leq m$.
\end{definition}

\begin{definition}[Causal Pareto front]
    The space of all Pareto-optimal intervention set-value pairs for a given $\mathcal{S} \subseteq \mathbb{P}(\mathbf{X})$ is called the \textit{causal Pareto set for $\mathcal{S}$}, denoted $\mathcal{P}^{\mathsf{c}}_s(\mathcal{S})$. The corresponding \textit{causal Pareto front for $\mathcal{S}$}, denoted $\mathcal{P}^{\mathsf{c} }_f(\mathcal{S})$, is the $m$-dimensional image of the causal Pareto set under the objectives $\mu_i$, $1 \leq i \leq m$.
\end{definition}

We define \textsc{mo-cbo} problems as identifying the causal Pareto front $\mathcal{P}_f^{\mathsf{c}}(\mathbb{P}(\mathbf{X}))$ over the set of all intervention sets, and we refer to $\mathcal{P}_f^{\mathsf{c}}(\mathbb{P}(\mathbf{X}))$ simply as the causal Pareto front. 

\subsection{Decomposition of \textsc{mo-cbo} Problems}
To navigate the discovery of Pareto-optimal intervention set-value pairs, we aim to simplify the search space.  

\begin{definition}[Local \textsc{mo-cbo} problem]\label{def:local_mocbo_problem}
    Let $\textbf{X}_s \in \mathbb{P}(\mathbf{X})$ be an intervention set. Then, the multi-objective optimization problem defined by the objective functions $\mu_i(\mathbf{X}_s,\ \cdot \ ): \mathcal{D}(\mathbf{X}_s) \rightarrow \mathbb{R}$, $\mathbf{x}_s \mapsto \mu_i(\mathbf{X}_s,\mathbf{x}_s)$, $1 \leq i \leq m$, is called the \textit{local \textsc{mo-cbo} problem w.r.t. $\mathbf{X}_s$}.
\end{definition}

\noindent
For the local \textsc{mo-cbo} problem w.r.t. $\mathbf{X}_s \in \mathbb{P}(\mathbf{X})$, its Pareto set shall be denoted as $\mathcal{P}_s^{\mathsf{l}} (\mathbf{X}_s)$ and the associated Pareto front as $\mathcal{P}_f^{\mathsf{l}}(\mathbf{X}_s)$. Each local \textsc{mo-cbo} problem corresponds to a standard multi-objective optimization task, solvable with existing methods. The following proposition decomposes \textsc{mo-cbo} problems into such local problems. The proof is given in \cref{appendix:mo_cbo_decomposition}.

\begin{proposition}\label{prop:mo_cbo.decomposition}
Given $\langle \mathcal{G}, \mathbf{Y},\mathbf{X}, \mathbf{C} \rangle$, let $\mathcal{S} \subseteq \mathbb{P}(\mathbf{X})$ be a non-empty collection of intervention sets. Then, it holds
\begin{equation}
    \mathcal{P}_f^{\mathsf{c}}(\mathcal{S}) \subseteq \bigcup_{s=1}^{|\mathcal{S}|} \mathcal{P}_f^{\mathsf{l}}(\mathbf{X}_s).
\end{equation}
\end{proposition}

\noindent
Therefore, discovering the causal Pareto front requires to identify Pareto-optimal solutions of the local \textsc{mo-cbo} problems with respect to all intervention sets $\mathbf{X}_s \in \mathbb{P}(\mathbf{X})$.
\section{Solving \textsc{mo-cbo} Problems}\label{sec:mo_cbo_solve}

In this section, we propose our methodology for solving \textsc{mo-cbo} problems, which is depicted in \cref{fig:methodology_overview}. In summary, we reduce the search space to a subset $\mathcal{S} \subseteq \mathbb{P}(\mathbf{X})$, solve the corresponding local \textsc{mo-cbo} problems to find their Pareto fronts, and extract only Pareto-optimal intervention-set value pairs to construct the causal Pareto front.

\subsection{Reducing the Search Space}\label{subsec:search_space_reduction}
The complexity of solving the local \textsc{mo-cbo} problems with $\mathcal{S} = \mathbb{P}(\mathbf{X})$ rises exponentially with the number of treatment variables, making this strategy impracticable for most tasks. This section proposes a superior method of exploiting the graph topology to identify a minimal subset $\mathcal{S} \subseteq \mathbb{P}(\mathbf{X})$ with $\mathcal{P}_f^{\textsf{c}}(\mathbb{P}(\mathbf{X})) = \mathcal{P}_f^{\textsf{c}}(\mathcal{S})$.
To this end, we generalize the results from \citet{NEURIPS2018_c0a271bc} to the multi-objective setting. All proofs and derivations are given \cref{appendix:search_space_reduction}. For now, we assume that there are no non-manipulative variables, i.e., $\mathbf{C} = \varnothing$.

First, we reduce the search space by disregarding intervention sets where some variables do not affect the targets:

\begin{definition}[Minimal intervention set] \label{def:causal_bandits.mis}
    A set $\mathbf{X}_s \in \mathbb{P}(\mathbf{X})$ is called a \textit{minimal intervention set} if, for every \textsc{scm} conforming to $\mathcal{G}$, there exists no subset $\mathbf{X}_s' \subset \mathbf{X}_s$ such that for all $\mathbf{x}_s \in \mathcal{D}(\mathbf{X}_s)$ it holds $\mu(\mathbf{X_s},\mathbf{x}_s) = \mu(\mathbf{X}_s',\mathbf{x}_s[\mathbf{X}_s'])$, for all $1 \leq i \leq m$.
\end{definition}

We denote the set of minimal intervention sets with $\mathbb{M}_{\mathcal{G},\mathbf{Y}}$. The following proposition characterizes such sets in a given causal graph $\mathcal{G}$. It is proven in \cref{subsec:causal_bandits_mis}.

\begin{proposition}\label{prop:causal_bandits.mis}
    $\mathbf{X}_s \in \mathbb{P}(\mathbf{X})$ is a minimal intervention set if and only if it holds $\mathbf{X}_s \subseteq \textnormal{an}(\mathbf{Y})_{\mathcal{G}_{\overline{\mathbf{X}}_s}}$.
\end{proposition}

Next, we adapt the notion of so called \textit{possibly-optimal minimal intervention sets} \cite{NEURIPS2018_c0a271bc} for Pareto-optimality. Intuitively said, a minimal intervention set is called possibly Pareto-optimal if it includes a Pareto-optimal intervention set-value pair whose outcome is unattainable with any other intervention set, for at least one \textsc{scm} conforming to $\mathcal{G}$.

\begin{definition}[Possibly Pareto-optimal minimal intervention set]\label{def:causal_bandits.pomis}
    A set $\mathbf{X}_s \in \mathbb{M}_{\mathcal{G},\mathbf{Y}}$ is called \textit{possibly Pareto-optimal} if, for at least one \textsc{scm} conforming to $\mathcal{G}$, there exists $\mathbf{x}_s \in \mathcal{D}(\mathbf{X}_s)$ such that $(\mathbf{X}_s,\mathbf{x}_s)$ is Pareto-optimal for $\mathbb{P}(\mathbf{X})$, and for no $\mathbf{X}_s' \in \mathbb{M}_{\mathcal{G},\mathbf{Y}} \backslash \mathbf{X}_s, \mathbf{x}_s' \in \mathcal{D}(\mathbf{X}_s')$ it holds $\mu(\mathbf{X}_s',\mathbf{x}_s') \leq \mu(\mathbf{X}_s,\mathbf{x}_s)$, for all $1\leq i \leq m$.
\end{definition}

We denote the set of possibly Pareto-optimal minimal intervention sets with $\mathbb{O}_{\mathcal{G},\mathbf{Y}}$. 
Next, we establish graph-theoretical criteria to identify such sets in a given causal graph. First, the proposition below, which we proof in \cref{subsec:causal_bandits_pomis}, considers a special case:

\begin{proposition}\label{prop:causal_bandits.pomis_no_confounders}
    If no $Y_i$ is confounded with $\textnormal{an}(Y_i)_{\mathcal{G}}$ via unobserved confounders, then $\textnormal{pa}(\mathbf{Y})_{\mathcal{G}}$ is the only possibly Pareto-optimal minimal intervention set.
\end{proposition}

To characterize possibly Pareto-optimal minimal intervention sets in arbitrary graphs, we extend the following two definitions from \citet{NEURIPS2018_c0a271bc} to the multi-objective setting. They aim to identify a region, starting from $\mathbf{Y}$, that is governed by unobserved confounders, along with its outside border that determines the realization of variables within the region.

\begin{definition}[Minimal unobserved confounders' territory]
    \label{def:mo_cbo.uc_territory}
    Let $\mathcal{H}=\mathcal{G}[\text{An}(\mathbf{Y})_{\mathcal{G}}]$. A set of variables $\mathbf{T}$ in $\mathcal{H}$, with $\mathbf{Y} \subseteq \mathbf{T}$, is called a \textit{UC-territory} for $\mathcal{G}$ w.r.t. $\mathbf{Y}$ if $\text{De}(\mathbf{T})_{\mathcal{H}}=\mathbf{T}$ and $\text{CC}(\mathbf{T})_{\mathcal{H}} = \mathbf{T}$. The UC-territory $\mathbf{T}$ is said to be \textit{minimal}, denoted $\mathbf{T} = \text{MUCT}(\mathcal{G},\mathbf{Y})$, if no $\mathbf{T}' \subset \mathbf{T}$ is a UC-territory.
\end{definition}

\begin{definition}[Interventional border]
    \label{def:mo_cbo.int_border}
    Let $\mathbf{T}=\text{MUCT}(\mathcal{G},\mathbf{Y})$. Then, $\mathbf{B} = \text{pa}(\mathbf{T})_{\mathcal{G}} \backslash \mathbf{T}$ is called the \textit{interventional border} for $\mathcal{G}$ w.r.t. $\mathbf{Y}$, denoted as $\text{IB}(\mathcal{G},\mathbf{Y})$.
\end{definition}

\begin{figure}[t]
    \centering
    \includegraphics{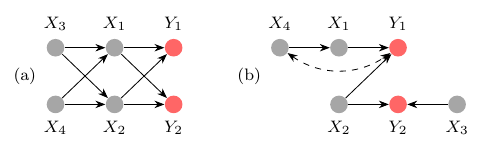}
    \caption{Two causal graphs with $\mathbf{X}=\{ X_1,X_2,X_3,X_4\}$, $\mathbf{Y}=\{Y_1,Y_2\}$. (a) No unobserved confounders. (b) An unobserved confounder between $X_4$ and $Y_1$ depicted with the dashed bi-directed edge.}
    \label{fig:synthtic_dags}
\end{figure}

\paragraph{Example} We illustrate these two concepts with the causal graphs from \cref{fig:synthtic_dags}. In \cref{fig:synthtic_dags} (a), there are no unobserved confounders and thus, it holds $\text{CC}(\mathbf{Y})_{\mathcal{G}} = \mathbf{Y}$ and $\text{De}(\mathbf{Y})_{\mathcal{G}} = \mathbf{Y}$. 
It follows $\text{MUCT}(\mathcal{G},\mathbf{Y}) = \{ Y_1,Y_2\}$ and $\text{IB}(\mathcal{G},\mathbf{Y})=\{X_1,X_2\}$. In \cref{fig:synthtic_dags} (b), we construct the minimal UC-territory, starting from $\mathbf{T}=\mathbf{Y}$,  as follows: 
Since $Y_1$ has an unobserved confounder with $X_4$, we update $\mathbf{T} = \text{CC}(\mathbf{Y})_{\mathcal{G}} = \{ Y_1,Y_2,X_4\}$, and thereafter add all the descendants of $X_4$, obtaining $\mathbf{T} = \{Y_1,Y_2,X_4,X_1\}$. 
Since there are no more unobserved confounders between $\mathbf{T}$ and $\text{An}(\mathbf{Y})_{\mathcal{G}} \backslash \mathbf{T}$, the minimal UC-territory has been found and is given by $\text{MUCT}(\mathcal{G},\mathbf{Y}) = \{ Y_1,Y_2, X_1, X_4\}$ along with $\text{IB}(\mathcal{G},\mathbf{Y})=\{X_2,X_3\}$. 

Interventional borders can fully characterize possibly Pareto-optimal minimal intervention sets, as seen in the following two results, both of which we prove in \cref{subsec:causal_bandits_pomis}.

\begin{proposition}
\label{prop:causal_bandits.intervenention_IB_pomis}
   $\textnormal{IB}(\mathcal{G}_{\overline{\mathbf{X}}_s}, \mathbf{Y})$ is a possibly Pareto-optimal minimal intervention set for any $\mathbf{X}_s \in \mathbb{P}(\mathbf{X})$.
\end{proposition}

\begin{theorem}\label{thm:causal_bandits}
    A set $\mathbf{X}_s \in \mathbb{P}(\mathbf{X})$ is a possibly Pareto-optimal minimal intervention set if and only if it holds $\textnormal{IB}(\mathcal{G}_{\overline{\mathbf{X}}_s}, \mathbf{Y}) = \mathbf{X}_s$.
\end{theorem}

The following corollary states that intervening on the interventional border of a set $\mathbf{X}_s \in \mathbb{P}(\mathbf{X})$ is at least as optimal as intervening on the set $\mathbf{X}_s$ itself.

\begin{corollary}\label{cor:causal_bandits}
    Let $\mathbf{X}_s \in \mathbb{P}(\mathbf{X})$ and $\mathbf{X}_s'=\textnormal{IB}(\mathcal{G}_{\overline{\mathbf{X}}_s}, \mathbf{Y})$. For any $\mathbf{x}_s \in \mathcal{D}(\mathbf{X}_s)$ there exist $\mathbf{x}_s' \in \mathcal{D}(\mathbf{X}_s')$ such that it holds $\mu(\mathbf{X}_s',\mathbf{x}_s') \leq \mu(\mathbf{X}_s,\mathbf{x}_s)$, for all $1\leq i \leq m$.
\end{corollary}

In the setting of \cref{cor:causal_bandits}, it is possible to construct an \textsc{scm} conforming to $\mathcal{G}$ such that strict inequality holds in at least one component. For such constructions, we refer to \cref{subsec:causal_bandits_pomis}. 

Finally, we show that it suffices to only consider possibly Pareto-optimal minimal intervention sets to solve \textsc{mo-cbo} problems. Due to the significance of this result, we also present its proof here.

\begin{proposition}\label{prop:pomis_suffice}
    It holds $\mathcal{P}_f^{\textsf{c}}(\mathbb{P}(\mathbf{X})) = \mathcal{P}_f^{\textsf{c}}(\mathbb{O}_{\mathcal{G},\mathbf{Y}})$.
\end{proposition}

\begin{proof}
    $\subseteq:$ Assume $\mathcal{P}_f^{\textsf{c}}(\mathbb{P}(\mathbf{X})) \not\subseteq \mathcal{P}_f^{\textsf{c}}(\mathbb{O}_{\mathcal{G},\mathbf{Y}})$. 
    Then, there exists $\mathbf{z} \in \mathbb{R}^m$, with $\mathbf{z}= \boldsymbol{\mu}(\mathbf{X}_s,\mathbf{x}_s)$ for some intervention set-value pair $(\mathbf{X}_s,\mathbf{x}_s)$, such that $\mathbf{z} \in \mathcal{P}_f^{\textsf{c}}(\mathbb{P}(\mathbf{X}))$ and $\mathbf{z} \not\in \mathcal{P}_f^{\textsf{c}}(\mathbb{O}_{\mathcal{G},\mathbf{Y}})$. 
    If $\mathbf{X}_s \in \mathbb{O}_{\mathcal{G},\mathbf{Y}}$, it follows that $(\mathbf{X}_s,\mathbf{x}_s)$ is not Pareto-optimal for $\mathbb{O}_{\mathcal{G},\mathbf{Y}}$, which is a contradiction since it is Pareto-optimal for $\mathbb{P}(\mathbf{X})$. 
    Conversely, if $\mathbf{X}_s \in \mathbb{P}(\mathbf{X}) \backslash  \mathbb{O}_{\mathcal{G},\mathbf{Y}}$, we set $\mathbf{X}_s' = \text{IB}(\mathcal{G}_{\overline{\mathbf{X}}_s},\mathbf{Y})$ and from \cref{cor:causal_bandits}, we infer that, for some \textsc{scm} conforming to $\mathcal{G},$ there exists $\mathbf{x}_s' \in \mathcal{D}(\mathbf{X}_s')$ with $\mu(\mathbf{X}_s',\mathbf{x}_s') \leq \mu(\mathbf{X}_s,\mathbf{x}_s)$, for all $1\leq i \leq m$, and $\mu(\mathbf{X}_s',\mathbf{x}_s') < \mu(\mathbf{X}_s,\mathbf{x}_s)$, for at least one $1\leq i \leq m$. This results in $\mathbf{z} \not\in \mathcal{P}_f^{\textsf{c}}(\mathbb{P}(\mathbf{X}))$, which is a contradiction. 

    $\supseteq:$ Assume $\mathcal{P}_f^{\textsf{c}}(\mathbb{O}_{\mathcal{G},\mathbf{Y}}) \not\subseteq  \mathcal{P}_f^{\textsf{c}}(\mathbb{P}(\mathbf{X})) $. Then, there exists $\mathbf{z} \in \mathbb{R}^m$, with $\mathbf{z}= \boldsymbol{\mu}(\mathbf{X}_s,\mathbf{x}_s)$, such that $\mathbf{z} \in \mathcal{P}_f^{\textsf{c}}(\mathbb{O}_{\mathcal{G},\mathbf{Y}})$ and $\mathbf{z} \not\in \mathcal{P}_f^{\textsf{c}}(\mathbb{P}(\mathbf{X}))$. There exists some $\mathbf{X}_s' \in \mathbb{P}(\mathbf{X}) \backslash  \mathbb{O}_{\mathcal{G},\mathbf{Y}}$, $\mathbf{x}_s \in \mathcal{D}(\mathbf{X}_s')$ such that $(\mathbf{X}_s',\mathbf{x}_s')$ is Pareto optimal and for which it holds $\mu(\mathbf{X}_s',\mathbf{x}_s') \leq \mu(\mathbf{X}_s,\mathbf{x}_s)$, for all $1\leq i \leq m$, and $\mu(\mathbf{X}_s',\mathbf{x}_s') < \mu(\mathbf{X}_s,\mathbf{x}_s)$, for at least one $1\leq i \leq m$. Since $\mathbf{X}_s'$ is not possibly Pareto-optimal, we infer from \cref{cor:causal_bandits} that for $\mathbf{X}_s'' = \text{IB}(\mathcal{G},\mathbf{Y})$ there exists $\mathbf{x}_s'' \in \mathcal{D}(\mathbf{X}_s'')$ such that $\mu(\mathbf{X}_s'',\mathbf{x}_s'') \leq \mu(\mathbf{X}_s',\mathbf{x}_s')$, for all $1\leq i \leq m$. Hence, it holds $\boldsymbol{\mu}(\mathbf{X}_s'',\mathbf{x}_s'') \in \mathcal{P}_f^{\textsf{c}}(\mathbb{P}(\mathbf{X}))$, which is a contradiction to $\mathbf{z} \in \mathcal{P}_f^{\textsf{c}}(\mathbb{O}_{\mathcal{G},\mathbf{Y}})$ since $\mathbf{X}_s'' \in \mathbb{O}_{\mathcal{G},\mathbf{Y}}$.
\end{proof}

Using \cref{prop:pomis_suffice}, we reduce the search space of \textsc{mo-cbo} problems to $\mathcal{S} = \mathbb{O}_{\mathcal{G},\mathbf{Y}}$.

\paragraph{Example} We illustrate the search space reduction with the causal graphs from \cref{fig:synthtic_dags}. Note that in both cases it holds $\mathbb{P}(\mathbf{X}) = 2^{|\mathbf{X}|} = 16$. In \cref{fig:synthtic_dags} (a), there are no unobserved confounders and it follows $\mathbb{O}_{\mathcal{G},\mathbf{Y}} = \text{pa}(\mathbf{Y})_{\mathcal{G}} = \{ X_1,X_2\}$. In \cref{fig:synthtic_dags} (b), the intervention sets $\{ X_1,X_2,X_3\}$ and  $\{X_2,X_3\}$ satisfy the condition from \cref{thm:causal_bandits}, and thus, $\mathbb{O}_{\mathcal{G},\mathbf{Y}} = \{ \{ X_2,X_3\}, \{ X_1,X_2,X_3\} \}$.

We now consider the more general case with $\mathbf{C} \neq \varnothing$, where non-manipulative variables can be present. The definitions for minimal intervention set and possibly Pareto-optimal minimal intervention set are a straightforward extension. 
\citet{lee2019structural} propose a projection $\mathcal{G} \rightarrow \mathcal{G}[\mathbf{V} \backslash \mathbf{C}]$ which preserves the distribution of the underlying \textsc{scm}. Given such a projection, we can identify the possibly Pareto-optimal minimal intervention sets in $\langle \mathcal{G},\mathbf{Y},\mathbf{X}, \mathbf{C} \rangle$ by applying \cref{thm:causal_bandits} to $\langle \mathcal{G}[\mathbf{V} \backslash \mathbf{C}],\mathbf{Y},\mathbf{X}\rangle$.

\subsection{Algorithm}\label{subsec:mo_cbo_algorithm}
We introduce the algorithm \textsc{Causal ParetoSelect}\footnote{The full implementation of our algorithm is available at \href{https://github.com/ShriyaBhatija/MO-CBO}{\texttt{https://github.com/ShriyaBhatija/MO-CBO}}.}, summarized in \cref{alg:cps}. It assumes a known causal graph $\langle \mathcal{G},\mathbf{Y}, \mathbf{X}, \mathbf{C} \rangle$, prior interventional data $\mathcal{D}^{I}$, and a set $\mathcal{S} \subseteq \mathbb{P}(\mathbf{X})$ that specifies which local problem to consider. The idea is to alternately solve the local \textsc{mo-cbo} problems using the batch multi-objective Bayesian optimization algorithm \textsc{dgemo} \cite{dgemo}.

\begin{algorithm}[tb]
   \caption{\textsc{Causal ParetoSelect}}
   \label{alg:cps}
\begin{algorithmic}
   \STATE {\bfseries Input:} $\langle \mathcal{G}, \mathbf{Y}, \mathbf{X}, \mathbf{C} \rangle$, $\mathcal{S} \subseteq \mathbb{P}(\mathbf{X})$, $\mathcal{D}^{I}$, batch size $B$, number of iterations $N$
   \STATE {\bfseries Output:} $\mathcal{P}_s^{\textsf{c}}(\mathcal{S})$, $\mathcal{P}_f^{\textsf{c}}(\mathcal{S})$
   \STATE Initialise the dataset $\mathcal{D}_0^{I}=\mathcal{D}^{I}$
   \FOR{$s=1$ {\bfseries to} $|\mathcal{S}|$}
   \STATE Fit surrogates $\tilde{\mu}_i(\mathbf{X_s}, \ \cdot \ )$ with $\mathcal{D}_0^{I}$, $i=1,\dots,m$
   \STATE Approximate $\mathcal{P}_s^{\textsf{l}}(\mathbf{X}_s)$ and $\mathcal{P}_f^{\textsf{l}}(\mathbf{X}_s)$ using $\tilde{\mu}_1,\dots,\tilde{\mu}_m$
   \ENDFOR
   \FOR{$n=1$ {\bfseries to} $N$}
   \FOR{$s=1$ {\bfseries to} $|\mathcal{S}|$}
   \STATE Select batch $\mathbf{B}_s = \{\mathbf{x}_{s}^{b}\}_{b=1}^B$ via \cref{eq:mo_cbo.dgemo_batch_selection}
   \ENDFOR
   \STATE Select batch $\mathbf{B}_{\hat{s}}$ from $\{ \mathbf{B}_1, \dots, \mathbf{B}_{|\mathcal{S}|} \}$ via \cref{eq:mo_cbo.batch_selection}
   \STATE Intervene on $\mathbf{X}_{\hat{s}}$ with $\mathbf{B}_{\hat{s}}$
   \STATE Augment $\mathcal{D}_n^{I} = \mathcal{D}_{n-1}^{I} \cup \{(\mathbf{X}_{\hat{s}},\mathbf{x}_{\hat{s}}^b),\boldsymbol{\mu}(\mathbf{X}_{\hat{s}}, \mathbf{x}_{\hat{s}}^b))\}_{b=1}^B$
   \STATE Update surrogates $\tilde{\mu}_i(\mathbf{X}_{\hat{s}}, \ \cdot \ )$ with $\mathcal{D}_{n}^{I}$, $i=1,\dots,m$
   \STATE Approximate $\mathcal{P}_s^{\textsf{l}}(\mathbf{X}_{\hat{s}})$ and $\mathcal{P}_f^{\textsf{l}}(\mathbf{X}_{\hat{s}})$ using $\tilde{\mu}_1,\dots,\tilde{\mu}_m$
   \ENDFOR
   \STATE Compute $\mathcal{P}_s^{\textsf{l}}(\mathbf{X}_s), \mathcal{P}_f^{\textsf{l}}(\mathbf{X}_s)$ from $\mathcal{D}_N^{I}$, $s=1,\dots, |\mathcal{S}|$
   \STATE Compute $\mathcal{P}^{\textsf{c}}_s(\mathcal{S})$ and $\mathcal{P}^{\textsf{c}}_f(\mathcal{S})$ 
\end{algorithmic}
\end{algorithm}

More specifically, \textsc{Causal ParetoSelect} operates as follows: For each local \textsc{mo-cbo} problem w.r.t. $\mathbf{X}_s \in \mathcal{S}$, we first fit the surrogate model to the objectives $\mu_i(\mathbf{X}_s, \ \cdot \ )$, $1\leq i\leq m$, via independent Gaussian processes as proposed by \textsc{dgemo}. Based on the means of the Gaussian process posteriors, we compute approximations of $\mathcal{P}_s^{\textsf{l}}(\mathbf{X}_s)$ and $\mathcal{P}_f^{\textsf{l}}(\mathbf{X}_s)$ utilizing the Pareto discovery approach from \textsc{dgemo}. After this initial step, at each iteration, the most promising intervention set is selected for batch evaluation. The dataset is augmented with the newly evaluated batch of samples. 
For the corresponding intervention set, we again update the surrogate model, as well as Pareto set and front approximations of the associated local problem. 
After completing all iterations, \textsc{Causal ParetoSelect} identifies the Pareto sets and fronts for each local \textsc{mo-cbo} problem using the collected objective function evaluations $\mathcal{D}_N^{I}$. 
Among those, $\mathcal{P}_s^{\textsf{c}}(\mathcal{S})$ and $\mathcal{P}_f^{\textsf{c}}(\mathcal{S})$ are determined by computing the Pareto-optimal points, which is justified by Proposition \ref{prop:mo_cbo.decomposition}.

\paragraph{Acquisition Function} 
We specify the batch selection strategy of our method in more detail. For a local \textsc{mo-cbo} problem w.r.t. $\mathbf{X}_s \in \mathcal{S}$, let $\mathcal{R}_1(\mathbf{X}_s),\dots,\mathcal{R}_K(\mathbf{X}_s) \subseteq \mathcal{D}(\mathbf{X}_s)$ denote the identified diversity regions from \textsc{dgemo}, discussed in \cref{subsec:prelim_mobo}. \textsc{Causal ParetoSelect} seeks to balance the exploration of the Pareto fronts associated to several different local \textsc{mo-cbo} problems. Evaluating all $\mathbf{B}_s$, $s=1,\dots,|\mathcal{S}|$, during a single iteration, is an inefficient strategy. Rather, at each iteration, we select the batch with the most promising hypervolume improvement. 
To this end, we introduce the term  \textit{relative hypervolume improvement}, which we define as
\begin{equation}
    \text{RHVI}(\boldsymbol{\mu}(\mathbf{X}_s,\mathbf{B}_s), \mathcal{P}_f^{\textsf{l}}(\mathbf{X_s})) = \frac{\text{HVI}(\boldsymbol{\mu}(\mathbf{X}_s,\mathbf{B}_s), \mathcal{P}^{{\textsf{l}}}_f(\mathbf{X_s}))}{\mathcal{H}(\mathcal{P}^{\textsf{l}}_f(\mathbf{X}_s))}.
\end{equation}
As the name suggests, relative hypervolume improvement is a normalized measure of improvement and therefore enables the assessment of batch evaluation across different intervention sets.
Given $\mathbf{B}_1,\dots,\mathbf{B}_{|\mathcal{S}|}$, we propose the following batch selection strategy for \textsc{Causal ParetoSelect}:
\begin{equation}\label{eq:mo_cbo.batch_selection}
    \mathbf{B}_{\hat{s}}= \underset{ \mathbf{B}_s \in \{\mathbf{B}_1,\dots,\mathbf{B}_{{|\mathcal{S}|}}\}}{\text{arg max }} \text{RHVI} (\boldsymbol{\mu}(\mathbf{X}_s,\mathbf{B}_s), \mathcal{P}_f^{\textsf{l}}(\mathbf{X_s})).
\end{equation}
Overall, the proposed batch selection is designed to alternately advance the Pareto fronts $\mathcal{P}_f^{\textsf{l}}(\mathbf{X}_1),\dots,\mathcal{P}_f^{\textsf{l}}(\mathbf{X}_{|\mathcal{S}|})$.

\section{Experiments}

We evaluate \textsc{Causal ParetoSelect} with $\mathcal{S} = \mathbb{O}_{\mathcal{G},\mathbf{Y}}$ on the causal graphs of \cref{fig:synthtic_dags} (a) (\textsc{synthetic-1}), \cref{fig:synthtic_dags} (b) (\textsc{synthetic-2}), and \cref{fig:health_dag} (\textsc{Health}), featuring both synthetic and real-world settings. The full description of the \textsc{scm}s can be found in \cref{appendix:experiments}. We assume an initial dataset $\mathcal{D}^{I} = \{ ((\mathbf{X}_s, \mathbf{x}_s^k), \boldsymbol{\mu}(\mathbf{X}_s, \mathbf{x}_s^k)) \}_{k=1,s=1}^{K,|\mathcal{S}|}$ with $K=5$ samples per intervention set. In each experiment, we aim to minimize the target variables. The batch size is set to \num{5}. For reproducibility, all experiments are run across 10 random seeds with varying initializations of $\mathcal{D}^{I}$. 

\begin{figure*}[t]
\vspace{-1cm}
\hspace{-0.6cm}
\includegraphics{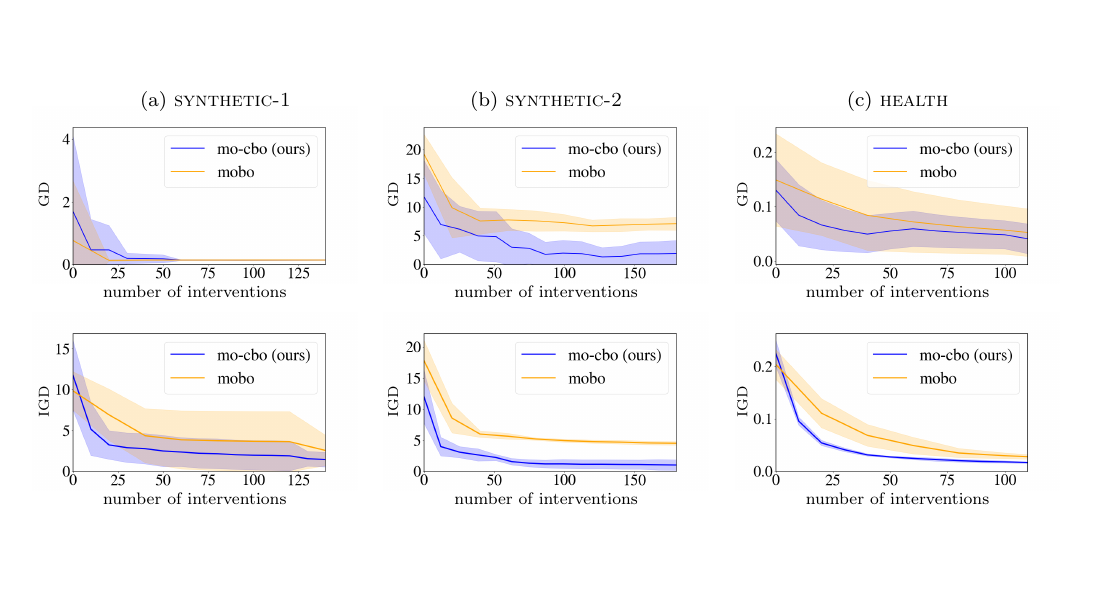}
\vspace{-1.7cm}
\caption{Convergence to the causal Pareto front across all experiments, measured by the generational distance (upper row) and inverted generational distance (lower row). Lower values indicate better approximations of the ground-truth. The number of interventions refers to the count of intervention variables that were intervened upon. Shaded areas represent $\pm$ standard deviation across \num{10} random seeds.
}
\label{fig:performance_metrics}
\end{figure*}

To the best of our knowledge, there exists no other multi-objective optimization method in the literature that can consider the causal structure $\langle \mathcal{G}, \mathbf{Y}, \mathbf{X}, \mathbf{C} \rangle$. 
Thus, our baseline method corresponds to the \textsc{mobo} algorithm \textsc{dgemo} \citep{dgemo}, applying it to all treatment variables simultaneously and neglecting available causal information.
Here, the objectives are $\mu_i(\mathbf{X}, \ \cdot \ ): \mathcal{D}(\mathbf{X}) \rightarrow \mathbb{R}$, $\mathbf{x} \mapsto \mu(\mathbf{X},\mathbf{x})$, $i=1,\dots,m$. 
We compare \textsc{Causal ParetoSelect} to the the baseline in regards to their approximations of $\mathcal{P}_f^{\textsf{c}}(\mathbb{P}(\mathbf{X}))$. 
To this end, we measure the accuracy of these approximations to the causal Pareto front using the generational distance (GD) as well as calculate the diversity of the solutions using the inverted generational distance (IGD) ~\cite{gd}.  
GD is defined as the average distance from any point in the approximated front to its closest point on the ground-truth front. Conversely,  IGD represents the average distance from any point in the ground-truth front to its closest point on the approximated front. 
The progressions of these performance metrics across all experiments are depicted in \cref{fig:performance_metrics}.

\subsection{Synthetic Problems}

\begin{figure}[h]
\centering
\vspace{-1.8cm}
\includegraphics{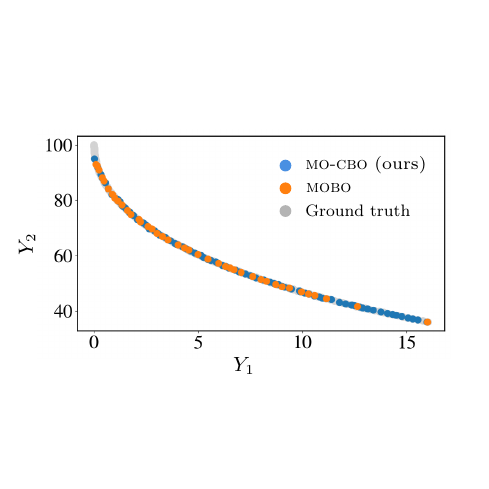}
\vspace{-2.3cm}
\caption{\textsc{synthetic-1}. Causal Pareto front approximations. Our method offers a higher coverage of the ground-truth causal Pareto front.}
\label{fig:synthetic1_fronts}
\end{figure}

\paragraph{\textsc{synthetic-1}} Here, it holds $\mathbb{O}_{\mathcal{G},\mathbf{Y}}= \{ \{X_1,X_2\} \}$ as discussed in \cref{subsec:search_space_reduction}.
The experimental results for \textsc{synthetic-1} are shown in \cref{fig:synthetic1_fronts} and demonstrate that the identified solutions from our method and \textsc{dgemo} closely align with the grount-truth front $\mathcal{P}_f^{\textsf{c}}(\mathbb{P}(\mathbf{X}))$. 
This alignment is further confirmed by the generational distance progressions in \cref{fig:performance_metrics} (a), which approach zero for both algorithms. Theoretically this result is expected, as $\mu(\mathbf{X},\mathbf{x}) = \mu(\mathbb{O}_{\mathcal{G},\mathbf{Y}},\mathbf{x}[\mathbb{O}_{\mathcal{G},\mathbf{Y}}])$ guarantees that the baseline contains Pareto-optimal solutions. Moreover, we observe that \textsc{Causal ParetoSelect} offers a better coverage of the front, a finding supported by its lower inverted generational distance in \cref{fig:performance_metrics} (a). This improvement stems from avoiding unnecessary interventions on $X_3$ and $X_4$, allowing for more exploratory interventions on $\mathbb{O}_{\mathcal{G},\mathbf{Y}}$ within the same number of interventions. 

\paragraph{\textsc{synthetic-2}} 
\begin{figure}[t]
\centering
\vspace{-1.8cm}
\includegraphics{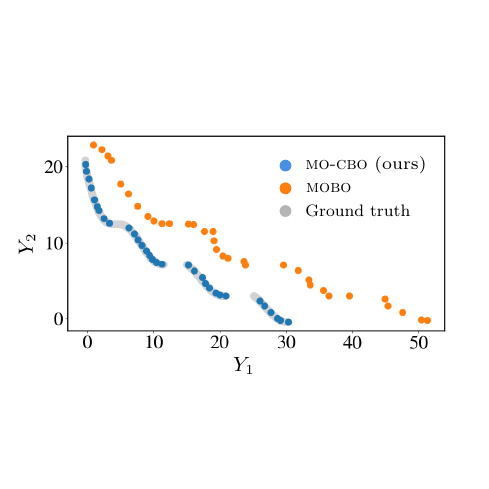}
\vspace{-2.6cm}
\caption{\textsc{synthetic-2}. Causal Pareto front approximations. By utilizing the causal relation, our approach tightly fits the ground-truth causal Pareto front.}
\label{fig:synthetic2_fronts}
\end{figure}
In this setting, an unobserved confounder exists between $Y_1$ and $X_4$, placing \textsc{synthetic-2} in the general case where hidden confounders may influence target variables through their ancestors. Consequently, it holds $\mathbb{O}_{\mathcal{G},\mathbf{Y}}=\{ \{X_2,X_3\}, \{X_1,X_2,X_3\} \}$. The experimental results, illustrated in \cref{fig:synthetic2_fronts}, demonstrate that while the baseline method fails to identify solutions on the causal Pareto front, \textsc{Causal ParetoSelect} does discover them. This finding is further reflected through the performance metrics in \cref{fig:performance_metrics}, which progress to significantly lower values for our method. 
Further experiments reveal that only interventions on $\{ X_2,X_3\}$ can yield solutions on the causal Pareto front. This can be explained as follows: The baseline strategy disrupts the causal path $X_4 \rightarrow X_1 \rightarrow Y_1$, letting the unobserved confounder to influence $Y_1$ without propagating through the aforementioned path. In contrast, our approach allows interventions on $\{X_2,X_3\}$, preserving this causal structure. This distinction is crucial as the structural assignment of $Y_1$ includes the term $-X_1 \cdot X_2 \cdot U/2$, with $U$ denoting the unobserved confounder (all structural equations are specified in \cref{subsec:experiments.synthetic-2}). Not intervening on $X_1$, causes this term to always be negative, yielding lower function values for $Y_1$. However, if we do intervene on $X_1$, it is positive with probability $0.5$, causing higher values for $Y_1$ in the averaged outcomes.

\subsection{Real-world Problem}

\begin{figure}[b]
\centering
\vspace{-1.8cm}
\includegraphics{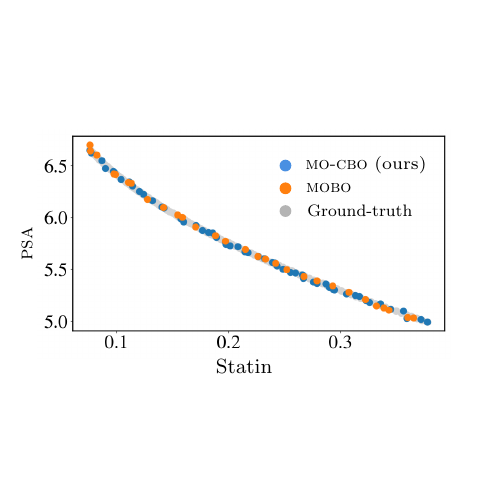}
\vspace{-2.6cm}
\caption{\textsc{Health}. Causal Pareto front approximations on a real-world healthcare application. Our approach yields a better coverage of the ground-truth.}
\label{fig:health_fronts}
\end{figure}

\paragraph{\textsc{health}} We revisit the causal graph from \cref{fig:health_dag}, which is based on real-world causal relationships in the healthcare setting \cite{ferro_healthcare}. 
For patients sensitive to Statin medication, one might aim to minimize both Statin usage and \textsc{psa} levels simultaneously. 
The possibly Pareto-optimal minimal intervention sets are $\mathbb{O}_{\mathcal{G},\mathbf{Y}} = \{ \{\textsc{bmi}, \text{Aspirin} \} \}$. 
The experimental results for \textsc{health}, depicted in \cref{fig:health_fronts}, show that both the baseline and \textsc{Causal ParetoSelect} identify solutions on the causal Pareto front.
The baseline method however, yields a sparser approximation, as confirmed by the progression of the inverted generational distance in \cref{fig:performance_metrics} (c). 
In contrast, \textsc{Causal ParetoSelect} discovers a larger number of solutions with the same number of interventions.

\section{Conclusion}
In this paper, we introduced \textsc{mo-cbo} as a new problem class in order to optimize multiple target variables within a known causal graph by sequentially performing interventions on the system. We proved that a \textsc{mo-cbo} problem can be decomposed into a collection of $|\mathbb{P}(\mathbf{X})| = 2^{|\mathbf{X}|}$ local problems, and solving it essentially corresponds to solving these local problems. To reduce the search space, we derived theoretical results that identify possibly Pareto-optimal minimal intervention sets in a given causal graph. We proved that these sets comprise a minimal collection of local problems that are guaranteed to contain the optimal solutions of any \textsc{mo-cbo} problem. Moreover, we introduced \textsc{Causal ParetoSelect} as an algorithm that iteratively selects and solves local \textsc{mo-cbo} problems in the reduced search space based on relative hypervolume improvement.

Our theoretical and empirical findings highlight two distinct cases: When no unobserved confounders exist between target variables and their ancestors, both \textsc{mo-cbo} and \textsc{mobo} can recover the ground-truth causal Pareto front. However, our approach demonstrates greater cost efficiency while constructing a more diverse set of solutions. In contrast, when unobserved confounders are present between targets and their ancestors, traditional \textsc{mobo} approaches can fail to approximate the ground truth, whereas \textsc{mo-cbo} demonstrates efficient discovery of Pareto-optimal solutions. This occurs because unobserved confounders can propagate effects through the causal graph, and naively disrupting these paths can lead to suboptimal solutions.

In our algorithm, the surrogate model assumes independent outcomes which may limit efficiency since it overlooks shared endogenous confounders. Future work could enhance cost effectiveness by integrating multi-task Gaussian processes to better capture shared information across treatment variables. Other directions for future research include the adaptation of existing \textsc{cbo} variants to the multi-objective case. For instance, combining dynamic \textsc{cbo} \citep{NEURIPS2021_577bcc91} with \textsc{mo-cbo} would lead to a \textsc{mo-cbo} variant that can handle time-dynamic causal models. As the field of causal decision-making continues to grow, we anticipate more progress in the development of multi-objective frameworks to address complex, real-world challenges.

\section*{Acknowledgments}
This work was financially supported by the Bavarian State Ministry of Education, Science and the Arts, the Alan Turing Institute’s studentship scheme, and the Engineering and Physical Sciences Research Council [EP/S023917/1].

\section*{Impact Statement}
This paper presents work whose goal is to advance the field of Machine Learning. 
There are many potential societal consequences of our work, none which we feel must be specifically highlighted here. We further emphasize that the health-related causal graph serves as a simplified illustrative example and, as such, carries no ethical concerns.

\bibliography{references}
\bibliographystyle{icml2025}

\newpage
\appendix
\onecolumn
\section{Decomposition of \textsc{mo-cbo} Problems}\label{appendix:mo_cbo_decomposition}

Recall the definition of the local \textsc{mo-cbo} problems.

\textbf{Definition \ref{def:local_mocbo_problem} \textnormal{(Local \textsc{mo-cbo} problem)}.}
Let $\textbf{X}_s \in \mathbb{P}(\mathbf{X})$ be an intervention set. Then, the multi-objective optimization problem defined by the objective functions $\mu_i(\mathbf{X}_s,\ \cdot \ ): \mathcal{D}(\mathbf{X}_s) \rightarrow \mathbb{R}$, $\mathbf{x}_s \mapsto \mu(\mathbf{X}_s,\mathbf{x}_s)$, $1 \leq i \leq m$, is called \textit{local \textsc{mo-cbo} problem w.r.t. $\mathbf{X}_s$}.

For the local \textsc{mo-cbo} problem w.r.t. $\mathbf{X}_s \in \mathbb{P}(\mathbf{X})$, we denote its Pareto set as $\mathcal{P}_s^{\textsf{l}}(\mathbf{X}_s)$ and the associated Pareto front as $\mathcal{P}_f^{\textsf{l}}(\mathbf{X}_s)$. 

\textbf{Proposition \ref{prop:mo_cbo.decomposition}.} \textit{Given $\langle \mathcal{G}, \mathbf{Y},\mathbf{X}, \mathbf{C} \rangle$, let $\mathcal{S} \subseteq \mathbb{P}(\mathbf{X})$ be a non-empty collection of intervention sets. Then, it holds}
\begin{equation}
    \mathcal{P}_f^{\textsf{c}}(\mathcal{S}) \subseteq \bigcup_{s=1}^{|\mathcal{S}|} \mathcal{P}_f^{\textsf{l}}(\mathbf{X}_s).
\end{equation}

\begin{proof}
    Assume for contradiction that $\mathcal{P}_f^{\textsf{c}}(\mathcal{S}) \not\subseteq \bigcup_{s=1}^{|\mathcal{S}|} \mathcal{P}_f^{\textsf{l}}(\mathbf{X}_s)$. Then, there exists some $\mathbf{z} \in \mathbb{R}^m$ such that $\mathbf{z} \in \mathcal{P}_f^{\textsf{c}}(\mathcal{S})$ and $\mathbf{z} \not\in \mathcal{P}_f^{\textsf{l}}(\mathbf{X}_{s})$ for all $s = 1, \dots, |\mathcal{S}|$. For some intervention set $\mathbf{X}_{s}' \in \mathcal{S}$ and intervention value $\mathbf{x}_s' \in \mathcal{D}(\mathbf{X}_s')$, it holds $\mathbf{z} = (\mu_1(\mathbf{X}_s',\mathbf{x}_s'),\dots,\mu_m(\mathbf{X}_s',\mathbf{x}_s'))$. Let $\mathcal{P}_s^{\textsf{l}}(\mathbf{X}_1),\dots,\mathcal{P}_s^{\textsf{l}}(\mathbf{X}_{|\mathcal{S}|})$ be the Pareto sets of the associated local \textsc{mo-cbo} problems w.r.t. $\mathbf{X}_1,\dots,\mathbf{X}_{|\mathcal{S}|}$, respectively. Since $\mathbf{z} \not\in \mathcal{P}_f^{\textsf{l}}(\mathbf{X}_s')$, it follows $\mathbf{x}_s' \not\in \mathcal{P}_s^{\textsf{l}}(\mathbf{X}_s')$, i.e. $\mathbf{x}_s'$ is not Pareto-optimal in the local \textsc{mo-cbo} problem w.r.t. $\mathbf{X}_s'$. Thus, there exists another intervention value $\mathbf{x}_s'' \in \mathcal{D}(\mathbf
    {X}_s')$ such that $\mu_i(\mathbf{X}_s',\mathbf{x}_s') \geq \mu_i(\mathbf{X}_s',\mathbf{x}_s'')$ for all $i$ and $\mu_i(\mathbf{X}_s',\mathbf{x}_s') > \mu_i(\mathbf{X}_s',\mathbf{x}_s'')$ for at least one $1 \leq i \leq m$. In other words, the intervention set-value pair $(\mathbf{X}_s',\mathbf{x}_s')$ is not Pareto-optimal for $\mathcal{S}$ since it is dominated by $(\mathbf{X}_s',\mathbf{x}_s'')$. Therefore, $\mathbf{z} \not\in \mathcal{P}_f^{\textsf{c}}(\mathcal{S})$ which is a contradiction.
\end{proof}

\section{Reducing the Search Space}\label{appendix:search_space_reduction}
\citet{NEURIPS2018_c0a271bc} leverage the graph topology of an \textsc{scm} to identify intervention sets that are redundant to consider in any optimisation scheme. Their formalism exploits the rules of do-calculus to identify invariances and partial-orders among intervention sets, in order to obtain those sets that could potentially yield optimal outcomes for a given graph. To take advantage of their ideas for this paper, the relevant concepts and their theoretical properties must be extended to accommodate multi-target settings, which will be the focus of this section.

Let $\langle \mathbf{V}, \mathbf{U}, \mathbf{F}, P(\textbf{U}) \rangle$ denote an \textsc{scm} and $\mathcal{G}$ its associated acyclic graph that encodes the underlying causal mechanisms. Recall that we assume $\mathbf{C} = \varnothing$, i.e., there are no non-manipulative variables. In this section, we require the notation $\mathbb{E}_{P(\mathbf{W} | \text{do}(\mathbf{X}_s = \mathbf{x}_s))}[\mathbf{W}] := \mathbb{E}[\mathbf{W} | \text{do}(\mathbf{X}_s = \mathbf{x}_s)]$ for sets $\mathbf{X}_s \subseteq \mathbf{X}$, $\mathbf{W} \subseteq \mathbf{V}$.

\subsection{Equivalence of Intervention Sets}\label{subsec:causal_bandits_mis}

As a first step, we establish invariances within $\mathbb{P}(\mathbf{X})$ in regards to the effects of intervention sets on the target variables. Recall the following definition from the main part of the paper.

\textbf{Definition \ref{def:causal_bandits.mis} \textnormal{(Minimal intervention set)}.} A set $\mathbf{X}_s \in \mathbb{P}(\mathbf{X})$ is called a \textit{minimal intervention set} if, for every \textsc{scm} conforming to $\mathcal{G}$, there exists no subset $\mathbf{X}_s' \subset \mathbf{X}_s$ such that for all $\mathbf{x}_s \in \mathcal{D}(\mathbf{X}_s)$ it holds $\mu(\mathbf{X_s},\mathbf{x}_s) = \mu(\mathbf{X}_s',\mathbf{x}_s[\mathbf{X}_s'])$, for all $1 \leq i \leq m$.

We denote the set of minimal intervention sets with $\mathbb{M}_{\mathcal{G},\mathbf{Y}}$. In other words, no subset of a minimal intervention set can achieve the same expected outcome on $\mathbf{Y}$. Intervention sets, that are not \textit{minimal} in the sense of Definition \ref{def:causal_bandits.mis}, are redundant to consider in any optimization task. 

\textbf{Proposition \ref{prop:causal_bandits.mis}.} 
\textit{$\mathbf{X}_s \in \mathbb{P}(\mathbf{X})$ is a minimal intervention set if and only if it holds $\mathbf{X}_s \subseteq \textnormal{an}(\mathbf{Y})_{\mathcal{G}_{\overline{\mathbf{X}}_s}}$.}

\begin{proof}
    (If) Let $\mathbf{x}_s \in \mathcal{D}(\mathbf{X}_s)$ be any intervention value. Assume that there is a subset $\mathbf{X}_s' \subset \mathbf{X}_s$ such that $\mathbb{E}[Y_i | \text{do}(\mathbf{X}_s = \mathbf{x}_s)] = \mathbb{E}[Y_i | \text{do}(\mathbf{X}_s' = \mathbf{x}_s[\mathbf{X}_s'])]$ for all $1 \leq i \leq m$. Consider an \textsc{scm} with real-valued variables where each $V \in \mathbf{V}$ is associated with its own binary exogenous variable $U_V$ with $P(U_V=1)=0.5$. Let the function of an endogenous variable be the sum of values of its parents. For the sake of contradiction, assume $\mathbf{X}_s \subseteq \text{an}(\mathbf{Y})_{\mathcal{G}_{\overline{\mathbf{X}}_s}}$. Then, there exists a directed path from $\mathbf{X}_s \backslash \mathbf{X}_s'$ to some $Y_i$ without passing $\mathbf{X}_s'$. Hence, setting $\mathbf{W} = \mathbf{X}_s \backslash \mathbf{X}_s'$ to the values \mbox{$\mathbf{w} = \mathbb{E}[\mathbf{W} | \text{do}(\mathbf{X}_s' = \mathbf{x}_s[\mathbf{X}_s'])] + 1$} yields \mbox{$\mathbb{E}[Y_i | \text{do}(\mathbf{W} = \mathbf{w}, \mathbf{X}_s' = \mathbf{x}_s[\mathbf{X}_s'])] >$} \mbox{$\mathbb{E}[Y_i | \text{do}(\mathbf{X}_s' = \mathbf{x}_s[\mathbf{X}_s'])]$}, contradicting the assumption.
    
    (Only if) Assume that $\mathbf{X}_s \not\subseteq \text{an}(\mathbf{Y})_{\mathcal{G}_{\overline{\mathbf{X}}_s}}$. Then, for $\mathbf{X}_s' = \mathbf{X}_s \cap \text{an}(\mathbf{Y})_{\mathcal{G}_{\overline{\mathbf{X}}_s}}$ it holds $\mathbf{X}_s' \subset \mathbf{X}_s$ and by the third rule of do-calculus, for every $\mathbf{x}_s \in \mathcal{D}(\mathbf{X}_s)$ it holds $\mathbb{E}[Y_i | \text{do}(\mathbf{X}_s = \mathbf{x}_s)] = \mathbb{E}[Y_i | \text{do}(\mathbf{X}_s' = \mathbf{x}_s[\mathbf{X}_s'])]$, $1 \leq i \leq m$. This is a contradiction because $\mathbf{X}_s$ was assumed to be a minimal intervention set.
\end{proof}

\subsection{Partial-Orders among Intervention Sets}\label{subsec:causal_bandits_pomis}

Recall the definition of possibly Pareto-optimal minimal intervention sets.

\textbf{Definition \ref{def:causal_bandits.pomis} \textnormal{(Possibly Pareto-optimal minimal intervention set)}.}
    A set $\mathbf{X}_s \in \mathbb{M}_{\mathcal{G},\mathbf{Y}}$ is called \textit{possibly Pareto-optimal} if, for at least one \textsc{scm} conforming to $\mathcal{G}$, there exists $\mathbf{x}_s \in \mathcal{D}(\mathbf{X}_s)$ such that $(\mathbf{X}_s,\mathbf{x}_s)$ is Pareto-optimal for $\mathbb{P}(\mathbf{X})$, and for no $\mathbf{X}_s' \in \mathbb{M}_{\mathcal{G},\mathbf{Y}} \backslash \mathbf{X}_s, \mathbf{x}_s' \in \mathcal{D}(\mathbf{X}_s')$ it holds $\mu(\mathbf{X}_s',\mathbf{x}_s') \leq \mu(\mathbf{X}_s,\mathbf{x}_s)$, for all $1\leq i \leq m$.

Characterizing such sets is the aim of this section. For simplicity, we first consider the special case in which $\mathcal{G}$ exhibits no unobserved confounders between $Y_i$ and any of its ancestors. 

\textbf{Proposition \ref{prop:causal_bandits.pomis_no_confounders}.} \textit{If no $Y_i$ is confounded with $\textnormal{an}(Y_i)_{\mathcal{G}}$ via unobserved confounders, then $\textnormal{pa}(\mathbf{Y})_{\mathcal{G}}$ is the only possibly Pareto-optimal minimal intervention set.}

\begin{proof}\label{proof:causal_bandits.pomis_no_confounders}
    Let $\mathbf{X}_s = \text{pa}(\mathbf{Y})_{\mathcal{G}}$, and let $\text{pa}(\mathbf{Y})_{\mathcal{G}} \neq \mathbf{X}_s'$ be another minimal intervention set with $\mathbf{x}_s' \in \mathcal{D}(\mathbf{X}_s')$. Define $\mathbf{Z}=\mathbf{X}_s \backslash (\mathbf{X}_s' \cap \mathbf{X}_s)$ and $\mathbf{W}=\mathbf{X}_s' \backslash (\mathbf{X}_s' \cap \mathbf{X}_s)$. Moreover, we choose an intervention value $\mathbf{x}_s^* \in \mathcal{D}(\mathbf{X}_s)$ such that it dominates $\mathbf{x}_s \in \mathcal{D}(\mathbf{X}_s)$ which is given by $\mathbf{x}_s[\mathbf{X}_s']=\mathbf{x}_s'[\mathbf{X}_s]$ and $\mathbf{x}_s[\mathbf{Z}] = \mathbb{E}[\mathbf{Z} | \text{do}(\mathbf{X}_s' = \mathbf{x}_s')]$. If $\mathbf{x}_s$ is non-dominated, define $\mathbf{x}_s^*=\mathbf{x}_s$. Then, for all $i=1,\dots,m$ it holds
    \begin{align}
        \mathbb{E} [Y_i | \text{do}(\mathbf{X}_s = \mathbf{x}_s^*)] & = \mathbb{E}[Y_i | \text{do}(\mathbf{X}_s \cap \mathbf{X}_s' = \mathbf{x}_s^*[\mathbf{X}_s'], \mathbf{Z} = \mathbf{x}_s^*[\mathbf{Z}])] \\
        &\leq \mathbb{E}[Y_i | \text{do}(\mathbf{X}_s \cap \mathbf{X}_s' = \mathbf{x}_s[\mathbf{X}_s'], \mathbf{Z} = \mathbf{x}_s[\mathbf{Z}])] \\
        & = \mathbb{E}[Y_i | \text{do}(\mathbf{X}_s \cap \mathbf{X}_s' = \mathbf{x}_s[\mathbf{X}_s'], \mathbf{Z} = \mathbf{x}_s[\mathbf{Z}], \mathbf{W}=\mathbf{x}_s'[\mathbf{W}])] \\
        & = \mathbb{E}[Y_i | \text{do}(\mathbf{X}_s \cap \mathbf{X}_s' = \mathbf{x}_s[\mathbf{X}_s'], \mathbf{W}=\mathbf{x}_s'[\mathbf{W}]), \mathbf{Z} = \mathbf{x}_s[\mathbf{Z}]] \\
        &= \mathbb{E}[Y_i | \text{do}(\mathbf{X}_s \cap \mathbf{X}_s' = \mathbf{x}_s[\mathbf{X}_s'], \mathbf{W}=\mathbf{x}_s'[\mathbf{W}])] \displaybreak[1] \\
        &= \mathbb{E}[Y_i | \text{do}(\mathbf{X}_s' = \mathbf{x}_s')],
    \end{align}
    where the inequality holds because $\mathbf{x}_s$ (weakly) dominates $\mathbf{x}_s^*$. Note that the second and third equalities are derived through the third and second rules of do-calculus, respectively. The second rule of do-calculus assumes that $Y_i$ is not confounded with $\text{an}(Y_i)_{\mathcal{G}}$ via unobserved confounders. For $\mathbf{X}_s = \text{pa}(\mathbf{Y})_{\mathcal{G}}$, it is possible to construct an \textsc{scm}, conforming to $\mathcal{G}$, such that strict inequality holds for some $Y_i$, see the proof of \autoref{prop:causal_bandits_IB}. This shows that $\text{pa}(\mathbf{Y})_{\mathcal{G}}$ is the only possibly Pareto-optimal minimal intervention set.
\end{proof}

We continue and study the more general case where unobserved confounders can be present between $Y_i$ and any of its ancestors. For this intent, we extend two existing concepts, called \textit{minimal unobserved-confounders’ territory} and \textit{interventional border} \citep{NEURIPS2018_c0a271bc}, to the multi-objective setting. Using these notions, we derive results which can fully characterize possibly Pareto-optimal minimal intervention sets in the aforementioned scenario.

\textbf{Definition \ref{def:mo_cbo.uc_territory} \textnormal{(Minimal unobserved confounders' territory)}.}
Let $\mathcal{H}=\mathcal{G}[\text{An}(\mathbf{Y})_{\mathcal{G}}]$. A set of variables $\mathbf{T}$ in $\mathcal{H}$, with $\mathbf{Y} \subseteq \mathbf{T}$, is called a \textit{UC-territory} for $\mathcal{G}$ w.r.t. $\mathbf{Y}$ if $\text{De}(\mathbf{T})_{\mathcal{H}}=\mathbf{T}$ and $\text{CC}(\mathbf{T})_{\mathcal{H}} = \mathbf{T}$. The UC-territory $\mathbf{T}$ is said to be \textit{minimal}, denoted $\mathbf{T} = \text{MUCT}(\mathcal{G},\mathbf{Y})$, if no $\mathbf{T}' \subset \mathbf{T}$ is a UC-territory.

A minimal UC-territory for $\mathcal{G}$ w.r.t. $\mathbf{Y}$ can be constructed by extending a set of variables, starting from $\mathbf{Y}$, and iteratively updating the set with the c-component and descendants of the set. More intuitively, it is the minimal subset of $\text{An}(\mathbf{Y})_{\mathcal{G}}$ that is governed by unobserved confounders, where at least one target $Y_i$ is adjacent to an unobserved confounder.

\textbf{Definition \ref{def:mo_cbo.int_border} \textnormal{(Interventional border)}.}
Let $\mathbf{T}=\text{MUCT}(\mathcal{G},\mathbf{Y})$. Then, $\mathbf{B} = \text{pa}(\mathbf{T})_{\mathcal{G}} \backslash \mathbf{T}$ is called the \textit{interventional border} for $\mathcal{G}$ w.r.t. $\mathbf{Y}$, denoted as $\text{IB}(\mathcal{G},\mathbf{Y})$.

We have already described these concepts in the main part. Before connecting the notion of minimal UC-territory and interventional border to possibly Pareto-optimal minimal intervention sets, we require the following proposition:

\begin{proposition}
    \label{prop:causal_bandits.subsumption}
    Let $\mathbf{T}$ be a minimal UC-territory and $\mathbf{B}$ an interventional border for $\mathcal{G}$ w.r.t. $\mathbf{Y}$. Let $\mathbf{X}_s \subseteq \mathbf{X}$ be an intervention set and $\mathbf{S} = (\mathbf{T} \cap \mathbf{X}_s) \cup \mathbf{B}$. Then, for any $\mathbf{x}_s \in \mathcal{D}(\mathbf{X}_s)$ there exists $\mathbf{s} \in \mathcal{D}(\mathbf{S})$ such that $\mathbb{E}[Y_i | \textnormal{do}(\mathbf{S}=\mathbf{s})] \leq \mathbb{E}[Y_i | \textnormal{do}(\mathbf{X}_s=\mathbf{x}_s)]$, for all $i=1,\dots,m$.
\end{proposition}

\begin{proof}
    (Case $\mathbf{B} \subseteq \mathbf{X}_s$) Let $\mathbf{x}_s \in \mathcal{D}(\mathbf{X}_s)$ be an intervention value. Then, by the third rule of do-calculus, it holds $\mathbb{E}[Y_i | \text{do}(\mathbf{X}_s=\mathbf{x}_s)]=\mathbb{E}[Y_i | \text{do}(\mathbf{X}_s \cap (\mathbf{T} \cup \mathbf{B}) =\mathbf{x}_s[\mathbf{T} \cup \mathbf{B}])]$, $1 \leq i \leq m$. Since $\mathbf{X}_s \cap (\mathbf{T} \cup \mathbf{B}) = \mathbf{S}$, by setting $\mathbf{s} = \mathbf{x}_s[\mathbf{T} \cup \mathbf{B}]$, it follows $\mathbb{E}[Y_i | \text{do}(\mathbf{X}_s=\mathbf{x}_s)] = \mathbb{E}[Y_i | \text{do}(\mathbf{S}=\mathbf{s})]$.
    \vspace{0.2cm}\\
    (Case $\mathbf{B} \not\subseteq \mathbf{X}_s$) Let $\mathbf{x}_s \in \mathcal{D}(\mathbf{X}_s)$ be an intervention value. We define $\mathbf{B}' = \mathbf{S} \backslash (\mathbf{X}_s \cap \mathbf{S}) = \mathbf{S} \backslash (\mathbf{X}_s \cap (\mathbf{T} \cup \mathbf{B})) = \mathbf{B} \backslash (\mathbf{X}_s \cap \mathbf{B})$ and $\mathbf{W} = \mathbf{X}_s \backslash (\mathbf{X}_s \cap \mathbf{S}) = \mathbf{X}_s \backslash (\mathbf{X}_s \cap (\mathbf{T} \cup \mathbf{B}))$. Moreover, let $\mathbf{s}^* \in \mathcal{D}(\mathbf{S})$ such that it dominates $\mathbf{s} \in \mathcal{D}(\mathbf{S})$, which is given by $\mathbf{s}[\mathbf{B}'] = \mathbb{E}[\mathbf{B}' | \text{do}(\mathbf{X}_s=\mathbf{x}_s)]$ and $\mathbf{s}[\mathbf{X}_s] = \mathbf{x}_s[\mathbf{T} \cup \mathbf{B}]$. If $\mathbf{s}$ is non-dominated, we set $\mathbf{s}^* = \mathbf{s}$. Then, for all $i=1,\dots,m$ it holds
    \begin{align}
        \mathbb{E}[Y_i | \text{do}(\mathbf{S}=\mathbf{s}^*)] &= \mathbb{E}[Y_i | \text{do}(\mathbf{X}_s \cap (\mathbf{T} \cup \mathbf{B}) =\mathbf{s}^*[\mathbf{X}_s], \mathbf{B}'=\mathbf{s}^*[\mathbf{B}'])]\\
        & \leq \mathbb{E}[Y_i | \text{do}(\mathbf{X}_s \cap (\mathbf{T} \cup \mathbf{B}) =\mathbf{s}[\mathbf{X}_s], \mathbf{B}'=\mathbf{s}[\mathbf{B}'])] \\
        &= \mathbb{E}[Y_i | \text{do}(\mathbf{X}_s \cap (\mathbf{T} \cup \mathbf{B}) =\mathbf{s}[\mathbf{X}_s], \mathbf{B}'=\mathbf{s}[\mathbf{B}'], \mathbf{W}=\mathbf{x}_s[\mathbf{W}])]\\
        &=\mathbb{E}[Y_i | \text{do}(\mathbf{X}_s \cap (\mathbf{T} \cup \mathbf{B}) =\mathbf{s}[\mathbf{X}_s], \mathbf{W}=\mathbf{x}_s[\mathbf{W}]), \mathbf{B}'=\mathbf{s}[\mathbf{B}']]\\
        &= \mathbb{E}[Y_i | \text{do}(\mathbf{X}_s \cap (\mathbf{T} \cup \mathbf{B}) =\mathbf{s}[\mathbf{X}_s], \mathbf{W}=\mathbf{x}_s[\mathbf{W}])]\\
        &= \mathbb{E}[Y_i | \text{do}(\mathbf{X}_s=\mathbf{x}_s)],
    \end{align}
    where the inequality holds because $\mathbf{s}$ is (weakly) dominated by $\mathbf{s}^*$. Furthermore, the second and third equalities are derived through the third and second rules of do-calculus, respectively.
\end{proof}

The following proposition is a building block for characterizing possibly Pareto-optimal minimal intervention sets via interventional borders. The proof is similar to the one given by \citet{NEURIPS2018_c0a271bc} 

\begin{proposition}
    \label{prop:causal_bandits_IB}
    $\textnormal{IB}(\mathcal{G},\mathbf{Y})$ is a possibly Pareto-optimal minimal intervention set.
\end{proposition}

\begin{proof}
    The intuition of this proof is to construct an \textsc{scm}, conforming to $\mathcal{G}$, for which the single best strategy involves intervening on $\text{IB}(\mathcal{G},\mathbf{Y})$. Let $\mathbf{T}$ and $\mathbf{B}$ denote $\text{MUCT}(\mathcal{G},\mathbf{Y})$ and $\text{IB}(\mathcal{G},\mathbf{Y})$, respectively. Every exogenous variable in $\mathbf{U}$ shall be a binary variable with its domain being $\{ 0, 1\}$. Let $\oplus$ denote the exclusive-or function and $\bigvee$ the logical OR operator.
    \vspace{0.2cm}\\
    (Case $\mathbf{T} = \mathbf{Y}$) 
    In this case, $\mathbf{B}$ corresponds to the parents of $\mathbf{Y}$. Therefore, no target variable $Y_i$ is confounded with $\text{an}(Y_i)_{\mathcal{G}}$ via unobserved confounders. Define an \textsc{scm} such that
    \begin{itemize}
        \item Each endogenous variable $V \in \mathbf{V}$ is influenced by an exogenous variable $U_V \in \mathbf{V}$;
        \item $f_{Y_i} = \bigvee \mathbf{u}^{Y_i} \oplus \bigvee \mathbf{pa}_{Y_i}$ with $P(\mathbf{U}^{Y_i}=0) \approx 1$, for all $i=1,\dots,m$;
        \item $f_{X} = (\bigoplus \mathbf{u}^X) \oplus (\bigoplus \mathbf{pa}_X)$ for $X \in \mathbf{X}$ and $P(U = 0) = 0.5$ for every $U \in \mathbf{U} \backslash (\bigcup_{i=1}^m \mathbf{U}^{Y_i})$. 
    \end{itemize}
    By the third rule of do-calculus and by taking conditional expectations, it holds 
    \begin{align}
        \mathbb{E}[Y_i | \text{do}(\mathbf{B}=0)] &= \mathbb{E}[Y_i | \text{do}(\text{pa}({Y_i})_{\mathcal{G}}=0)] \\
        &= \mathbb{E}[Y_i | \text{do}(\text{pa}({Y_i})_{\mathcal{G}}=0), \mathbf{U}^{Y_i}\neq0] P(\mathbf{U}^{Y_i}\neq0) + \mathbb{E}[Y_i | \text{do}(\text{pa}({Y_i})_{\mathcal{G}}=0), \mathbf{U}^{Y_i}=0] P(\mathbf{U}^{Y_i}=0)\\
        &\approx 0
    \end{align}
    for every $1 \leq i \leq m$. Meanwhile, all other interventions yield expectations greater than or equal to 0.5 in at least one component. Therefore, $\mathbf{B}$ is a possibly Pareto-optimal minimal intervention set.
 
    (Case $\mathbf{T} \subset \mathbf{Y}$) In this case, at least one target variable $Y_i$ has an unobserved confounder with its ancestors. As a first step, it will be shown that there exists an \textsc{scm}, conforming to $\mathcal{H} = \mathcal{G}[\mathbf{T} \cup \mathbf{B}]$, where the intervention $\text{do}(\mathbf{B} = 0)$ is the single best strategy. To achieve this, we first define individual \textsc{scm}s for each unobserved confounder in $\mathcal{H}[\mathbf{T}]$, and merge them into a single \textsc{scm} where $\text{do}(\mathbf{B} = 0)$ is indeed the best strategy. Let $\mathbf{U}' = \{ U_j\}_{j=1}^k$ be the set of unobserved confounders in $\mathcal{H}[\mathbf{T}]$.

    Given $U_j \in \mathbf{U}'$, let $B^{(j)}$ and $R^{(j)}$ denote its two children. We define an \textsc{scm} $\mathcal{M}_j$, where the graph structure is given by
 
    \begin{equation}
        \mathcal{H}_j = \mathcal{H} \left[ \text{De}\left( \left\{ B^{(j)}, R^{(j)}\right\}\right)_{\mathcal{H}} \cup \left( \mathbf{B} \cap \text{pa}\left(\text{De} \left( \left\{ B^{(j)}, R^{(j)}\right\} \right)_{\mathcal{H}} \right) \right)\right],
    \end{equation}
    and all bidirected edges, except for $U_j$, are removed. In order to set the structural equations for variables in $\mathcal{H}_j$, the vertices will be labelled via colour coding: Let vertices in $\text{De}\left(B^{(j)}\right)_{\mathcal{H}} \backslash \text{De}\left(R^{(j)}\right)_{\mathcal{H}}$ be labelled as \textcolor{blue}{blue}, $\text{De}\left(R^{(j)}\right)_{\mathcal{H}} \backslash \text{De}\left(B^{(j)}\right)_{\mathcal{H}}$ as \textcolor{red}{red}, and $\text{De}\left(B^{(j)}\right)_{\mathcal{H}} \cap \text{De}\left(R^{(j)}\right)_{\mathcal{H}}$ as \textcolor{purple}{purple}. All target variables are coloured as \textcolor{purple}{purple} as well. Moreover, $B^{(j)}$ and $R^{(j)}$ shall perceive $U_j$ as a parent coloured as \textcolor{blue}{blue} with value $U_j$ and \textcolor{red}{red} with value $1-U_j$, respectively. The \textcolor{blue}{blue}-, \textcolor{red}{red}- and \textcolor{purple}{purple}-coloured variables are set to 3 if any of their parents in $\mathbf{B}$ is not 0. Otherwise, their values are determined as follows. For every \textcolor{blue}{blue} and \textcolor{red}{red} vertex, the associated structural equation returns the common value of its parents of the same colour and returns 3 if coloured parents' values are not homogeneous. For every \textcolor{purple}{purple} vertex, its corresponding equation returns 2 if every \textcolor{blue}{blue}, \textcolor{red}{red} and \textcolor{purple}{purple} parent is 0,1, and 2, respectively, and returns 1 if 1,0,1, respectively.

    Next, the \textsc{scm}s $\mathcal{M}_1,\dots,\mathcal{M}_k$ will be merged into one single \textsc{scm}, that conforms to $\mathcal{H}$, and for which $\text{do}(\mathbf{B}=0)$ is the single best intervention. Note that in $\mathcal{M}_j$ all variables can be represented with just two bits. To construct a unified \textsc{scm}, variables in $\mathbf{T}$ are represented with $2k$ bits, where $\mathcal{M}_j$ takes the $2j-1^{\text{th}}$ and $2j^{\text{th}}$ bits. Every target variable $Y_i$ is represented as a sequence of bits and binarised as follows. $Y_i$ is set to $0$ if its $2j-1^{\text{th}}$ and $2j^{\text{th}}$ bits are 00, 01 or 10 for every $1 \leq j \leq k$, and $1$ otherwise. Let $P(U_j=1) = 0.5$ for $U_j \in \mathbf{U}'$. Therefore, it holds $Y_i = 0$ if $\text{do}(\mathbf{B}=0)$ and $Y_i=1$ if $\text{do}(\mathbf{B}\neq 0)$.
    If any variable in $\mathbf{T}$ is intervened, then at least one \textsc{scm} $\mathcal{M}_j$ will be disrupted, resulting in an expectation larger than or equal to 0.5 for at least one target variable.
    In the multi-target setting, it may happen that some target variables do not occur in any of the $\mathcal{M}_j$'s. This happens if a target $Y_i$ has no parents in  $\mathbf{T}$, but only in $\mathbf{B}$. For all such $Y_i$'s, we set $f_{Y_i} = \mathbf{u}^{Y_i} \oplus \bigvee \mathbf{pa}_{Y_i}$ with $P(\mathbf{U}^{Y_i} = 0) \approx 1$. As such, the newly constructed \textsc{scm} enforces $\mathbb{E}[Y_i | \text{do}(\mathbf{B}=0)] \approx 0$. Meanwhile, all other interventions yield expectations greater than or equal to 0.5
    \vspace{0.2cm}\\
    As a last step, the previously defined \textsc{scm} for $\mathcal{H} = \mathcal{G}[\mathbf{T} \cup \mathbf{B}]$, will be extended to an \textsc{scm} for $\mathcal{G}$. However, we can ignore joint probability distributions for any exogenous variables only affecting endogenous variables outside of $\mathcal{H}$. Setting structural equations for endogenous variables outside of $\mathcal{H}$ is redundant as well. For $V \in \text{An}(\mathbf{Y})_{\mathcal{G}} \backslash \mathbf{T}$, we define the structural equations as $f_V = (\bigoplus \mathbf{u}^V) \oplus (\bigoplus \mathbf{pa}_V)$. For $U \in \mathbf{U} \backslash \mathbf{U}'$, we set $P(U=0)=0.5$ if $U$'s child(ren) is disjoint to $\mathbf{T}$, and $P(U=0)\approx 1$ otherwise. Note that $\text{do}(\mathbf{B}=0)$ is still the single optimal intervention. Therefore, $\mathbf{B}$ is a possibly Pareto-optimal minimal intervention set. 
\end{proof}

\begin{figure}[t]
    \centering
    \includegraphics[width=0.6\linewidth]{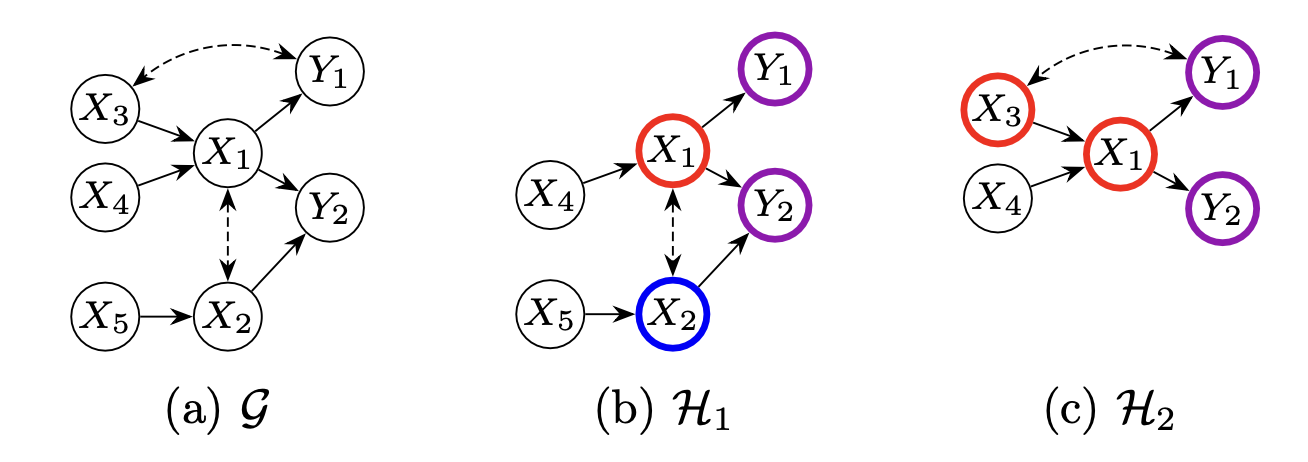}
    \caption{Original causal graph $\mathcal{G}$ and its color-coded subgraphs for each unobserved confounder.}
    \label{fig:causal_bandits.IB}
\end{figure}
\begin{figure}[t]
    \centering
    \includegraphics[width=0.98\linewidth]{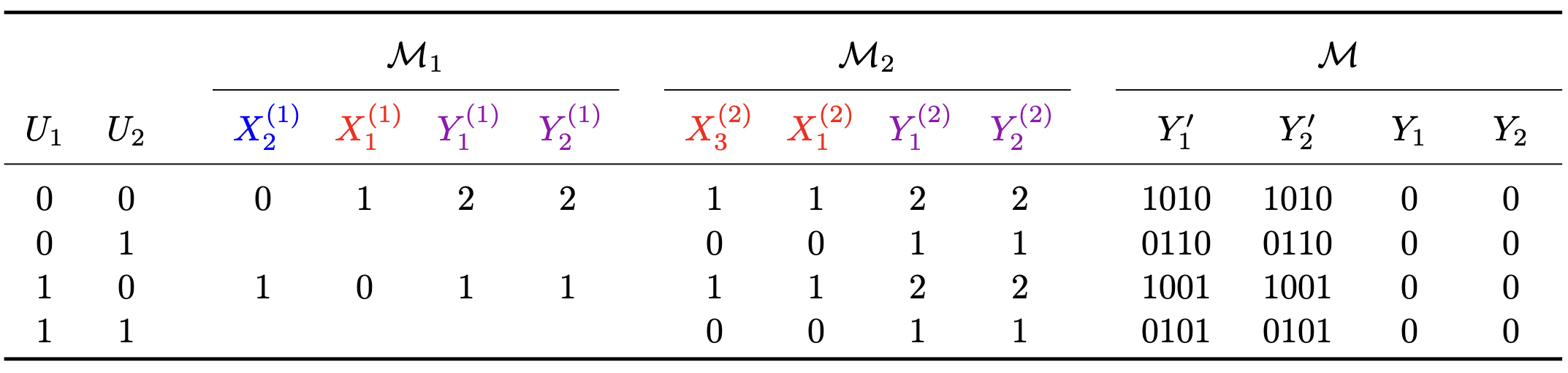}
    \caption{Values for $\mathcal{M}_1$, $\mathcal{M}_2$ and $\mathcal{M}$ given $X_4= X_5=0$. The target variables are shown as bit sequences, $Y_1'$ and $Y_2'$, as well as binary values, $Y_1$ and $Y_2$.}
    \label{tab:causal_bandits_IB}
\end{figure} 

In order to illustrate the construction of an \textsc{scm} where $\text{do}(\text{IB}(\mathcal{G}, \mathbf{Y}) = 0)$ is the single best strategy, consider \cref{fig:causal_bandits.IB}, showing an exemplary graph and its colour-coded subgraphs, $\mathcal{H}_1$ and $\mathcal{H}_2$, for each unobserved confounder. \cref{tab:causal_bandits_IB} presents the associated values for $\mathcal{M}_1$ and $\mathcal{M}_2$, as well as values for the target variables in the final \textsc{scm} $\mathcal{M}$. The next proposition generalizes the previous one.

\textbf{Proposition \ref{prop:causal_bandits.intervenention_IB_pomis}.}
\textit{$\textnormal{IB}(\mathcal{G}_{\overline{\mathbf{X}}_s}, \mathbf{Y})$ is a possibly Pareto-optimal minimal intervention set for any $\mathbf{X}_s \in \mathbb{P}(\mathbf{X})$.}

\begin{proof}
  Let $\mathbf{X}_s$ be an intervention set. Let us denote $\mathbf{T} = \text{MUCT}(\mathcal{G}_{\overline{\mathbf{X}}_s},\mathbf{Y})$, $\mathbf{B} = \text{IB}(\mathcal{G}_{\overline{\mathbf{X}}_s},\mathbf{Y})$ and $\mathbf{T}_0 = \text{MUCT}(\mathcal{G},\mathbf{Y})$. Using the strategy from \autoref{prop:causal_bandits.intervenention_IB_pomis}, we construct an \textsc{scm} for $\mathcal{G}[\mathbf{T} \cup \mathbf{B}]$ while ignoring unobserved confounders between $\mathbf{T}$ and $\mathbf{T}_0 \backslash \mathbf{T}$. Let $\mathbf{U}'$ be the set of such unobserved confounders. Now, the \textsc{scm} needs to be modified to ensure that $\text{do}(\mathbf{B}=0)$ is the single best intervention. Every $U \in \mathbf{U}'$ shall flip (i.e., $0 \xleftrightarrow{} 1$) the value of its endogenous child in $\mathbf{T}$ whenever $U=1$. Let $P(U=0) \approx 1$, so that it holds $\mathbb{E}[Y_i | \text{do}(\mathbf{B}=0)] \approx 0$. Intervening on $\mathbf{B} \neq 0$ or on any variable in $\mathbf{T}$ results in expectations around 0.5 or above.
\end{proof}

Notably, Proposition \ref{prop:causal_bandits.intervenention_IB_pomis} extends Proposition \ref{prop:causal_bandits_IB} when $\mathbf{X}_s \neq \varnothing$. Note that, by iterating over all intervention sets $\mathbf{X}_s \in \mathbb{P}(\mathbf{X})$, we can discover possibly Pareto-optimal minimal intervention sets in a given graph. The following theorem is an extension of the main result by \citet{NEURIPS2018_c0a271bc} to the scenario where multiple target variables are present. It shows that the aforementioned strategy suffices to find not some, but all, such sets.

\textbf{Theorem \ref{thm:causal_bandits}.}
    \textit{A set $\mathbf{X}_s$ is a possibly Pareto-optimal minimal intervention set if and only if it holds $\textnormal{IB}(\mathcal{G}_{\overline{\mathbf{X}}_s}, \mathbf{Y}) = \mathbf{X}_s$.}

\begin{proof}
    (If) This is a special case of Proposition \ref{prop:causal_bandits.intervenention_IB_pomis}.
    \vspace{0.2cm}\\
    (Only if) Let $\mathbf{X}_s$ be a minimal intervention set and $\mathbf{x}_s \in \mathcal{D}(\mathbf{X}_s)$ an intervention value. Denote $\mathbf{T}=\text{MUCT}(\mathcal{G}_{\overline{\mathbf{X}}_s},\mathbf{Y})$, $\mathbf{B}=\text{IB}(\mathcal{G}_{\overline{\mathbf{X}}_s},\mathbf{Y})$, $\mathbf{T}_0=\text{MUCT}(\mathcal{G},\mathbf{Y})$ and $\mathbf{B}_0=\text{IB}(\mathcal{G},\mathbf{Y})$. From \autoref{prop:causal_bandits.subsumption}, we know that no \textsc{pomis} intersects with $\text{An}(\mathbf{B}_0)_{\mathcal{G}} \backslash \mathbf{B}_0$ and thus, it is possible to conclude $\mathbf{X}_s \subseteq \mathbf{T}_0 \cup \mathbf{B}_0 \backslash \mathbf{Y}$. Note that it holds $\mathbf{X}_s \subseteq \text{An}(\mathbf{B})_{\mathcal{G}}$ since otherwise it would follow $\mathbf{X}_s \cap \mathbf{T} \neq \emptyset$, which contradicts that $\mathbf{X}_s$ is neither a descendant of some variable nor confounded in $\mathcal{G}_{\overline{\mathbf{X}}_s}$. Let $\mathbf{B}' = \mathbf{B} \backslash (\mathbf{X}_s \cap \mathbf{B})$ and $\mathbf{W} = \mathbf{X}_s \backslash (\mathbf{X}_s \cap \mathbf{B})$. Moreover, we define an intervention value $\mathbf{b}^* \in \mathcal{D}(\mathbf{B})$ such that it dominates $\mathbf{b} \in \mathcal{D}(\mathbf{B})$, which is given by $\mathbf{b}[\mathbf{B}'] = \mathbb{E}[\mathbf{B}' | \text{do}(\mathbf{X}_s= \mathbf{x}_s)]$ and $\mathbf{b}[\mathbf{X}_s] = \textbf{x}_s[\mathbf{B}]$. If $\mathbf{b}$ is non-dominated, we set $\mathbf{b}^*=\mathbf{b}$. Then, for all $i=1,\dots,m$, it holds
    \begin{align}
        \mathbb{E}[Y_i | \text{do}(\mathbf{B}=\mathbf{b}^*)] &=  \mathbb{E}[Y_i | \text{do}(\mathbf{B} \cap \mathbf{X}_s = \mathbf{b}^*[\mathbf{X}_s], \mathbf{B}' = \mathbf{b}^*[\mathbf{B}'])]  \\
        &\geq  \mathbb{E}[Y_i | \text{do}(\mathbf{B} \cap \mathbf{X}_s = \mathbf{b}[\mathbf{X}_s], \mathbf{B}' = \mathbf{b}[\mathbf{B}'])] \\
        &=  \mathbb{E}[Y_i | \text{do}(\mathbf{B} \cap \mathbf{X}_s = \mathbf{b}[\mathbf{X}_s], \mathbf{B}' = \mathbf{b}[\mathbf{B}'], \mathbf{W}=\mathbf{x}_s[\mathbf{W}])] \\
        &=  \mathbb{E}[Y_i | \text{do}(\mathbf{B} \cap \mathbf{X}_s = \mathbf{b}[\mathbf{X}_s], \mathbf{W}=\mathbf{x}_s[\mathbf{W}]), \mathbf{B}' = \mathbf{b}[\mathbf{B}']]\\
        &=  \mathbb{E}[Y_i | \text{do}(\mathbf{B} \cap \mathbf{X}_s = \mathbf{b}[\mathbf{X}_s], \mathbf{W}=\mathbf{x}_s[\mathbf{W}])]\\
        &= \mathbb{E}[Y_i | \text{do}(\mathbf{X}_s = \mathbf{x}_s)],
    \end{align}
    where the inequality holds because $\mathbf{b}$ is (weakly) dominated by $\mathbf{b}^*$. Furthermore, the second and third equalities are derived through the third and second rules of do-calculus, repectively.
\end{proof}

\autoref{thm:causal_bandits} provides a necessary and sufficient condition for a set of variables to be a possibly Pareto-optimal minimal intervention set. The proof of the theorem gives the following corollary:

\textbf{Corollary \ref{cor:causal_bandits}.}
    \textit{Let $\mathbf{X}_s \in \mathbb{P}(\mathbf{X})$ and $\mathbf{X}_s'=\textnormal{IB}(\mathcal{G}_{\overline{\mathbf{X}}_s}, \mathbf{Y})$. For any $\mathbf{x}_s \in \mathcal{D}(\mathbf{X}_s)$ there exist $\mathbf{x}_s' \in \mathcal{D}(\mathbf{X}_s')$ such that it holds $\mu(\mathbf{X}_s',\mathbf{x}_s') \leq \mu(\mathbf{X}_s,\mathbf{x}_s)$, for all $1\leq i \leq m$.}

\section{Experiments}\label{appendix:experiments}

\subsection{\textsc{synthetic-1}}\label{subsec:experiments.synthetic-1}
\begin{figure}[t]
    \centering
    \vspace{-1.5cm}
    \includegraphics{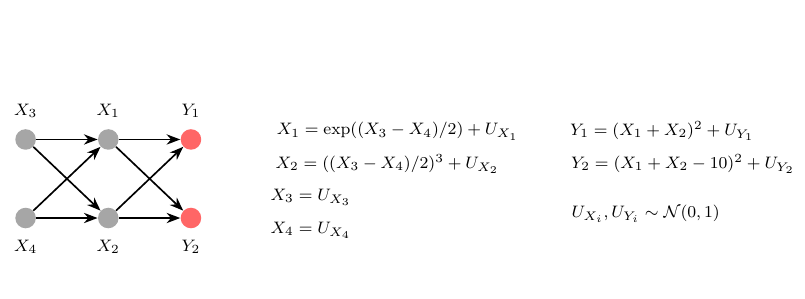}
    \vspace{-0.7cm}
    \caption{\textsc{synthetic-1}. An \textsc{scm} consisting of four treatment and two output variables, depicted with grey and red nodes, respectively. There are no unobserved confounders.}
    \label{fig:experiments.synthetic_1.scm}
\end{figure}

We introduce the first synthetic \textsc{mo-cbo} problem in our experimental study, referred to as \textsc{synthetic-1}, which is defined by the causal graph $\mathcal{G}$ and associated structural assignments presented in \autoref{fig:experiments.synthetic_1.scm}. The interventional domains are specified as $\mathcal{D}(X_1),\mathcal{D}(X_2) = [-1,2]$ and $\mathcal{D}(X_3),\mathcal{D}(X_4) = [-1,1]$. Moreover, all exogenous variables follow the standard normal distribution, and there are no unobserved confounders. 

\subsection{\textsc{synthetic}-2}\label{subsec:experiments.synthetic-2}
\begin{figure}[h]
    \centering
    \vspace{-1cm}
    \includegraphics{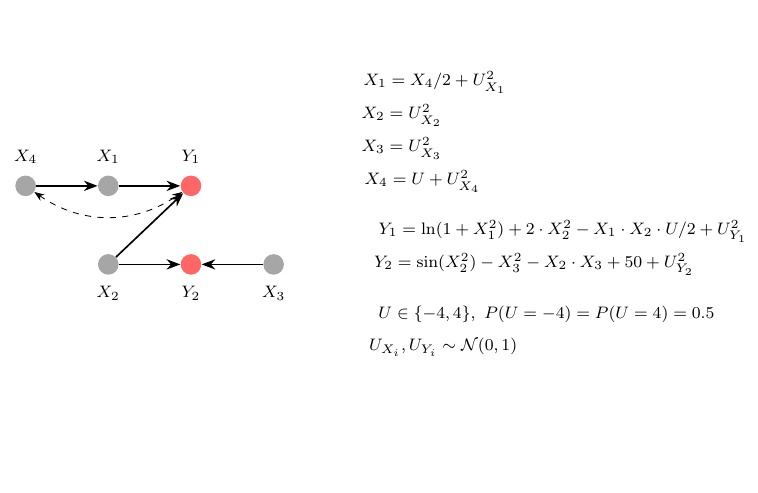}
    \vspace{-2.5cm}
    \caption{\textsc{synthetic-2}. An \textsc{scm} consisting of four treatment and two output variables, depicted with grey and red nodes, respectively. It includes an unobserved confounder, denoted via the dashed bi-directed edge, affecting one output and its ancestor.}
    \label{fig:experiments.synthetic-2.scm}
\end{figure}

\textsc{synthetic-2} is the next \textsc{mo-cbo} problem of our experimental study, defined by the causal graph $\mathcal{G}$ and associated structural equations in \autoref{fig:experiments.synthetic-2.scm}. The interventional domains are $\mathcal{D}(X_1) = [-2,5]$, $\mathcal{D}(X_4) = [-4,5]$ and $\mathcal{D}(X_i) = [0,5]$ for $i=1,2$. Moreover, the exogenous variables $U_{X_i}, U_{Y_i}$ follow a Gaussian distribution, and there is an unobserved confounder $U$ influencing the target variable $Y_1$ and its ancestor $X_4$.

\subsection{\textsc{health}}
\begin{figure}[t!]
    \centering
    \vspace{-17cm}
    \includegraphics{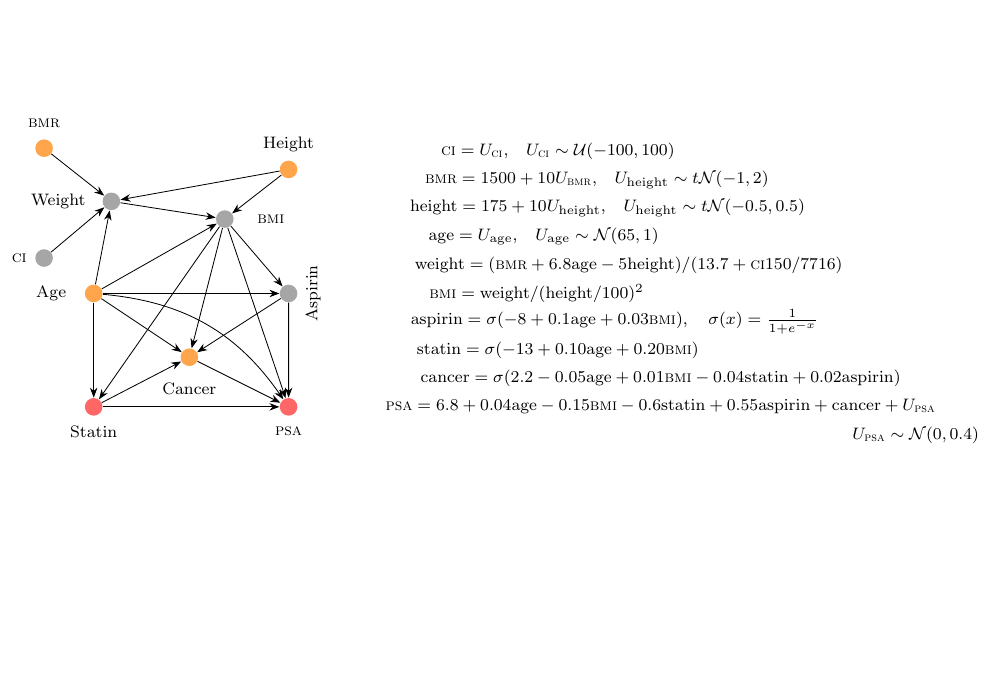}
    \vspace{-4.2cm}
    \caption{\textsc{Health}. An \textsc{scm} with relations between variables such as age, \textsc{bmi}, aspirin and statin usage, and their effects on \textsc{psa} levels \citep{pmlr-v216-gultchin23a}. $\mathcal{U}(\cdot,\cdot)$ denotes a uniform distribution and $t\mathcal{N}(a,b)$ a standard Gaussian distribution truncated between $a$ and $b$. Red, grey, and orange nodes depict target, manipulative, and non-manipulative variables, respectively.}
    \label{fig:experiments.healthcare.scm}
\end{figure}

The \textsc{mo-cbo} problem \textsc{Health} is defined by the causal graph and structural equations in \cref{fig:experiments.healthcare.scm}. This model originates from previous works of \citet{ferro_healthcare}, and is based on real-world causal relationships. It captures prostate-specific-antigen (\textsc{psa}) levels in causal relation to its risk factors, such as \textsc{bmi}, calorie intake (\textsc{ci}) and aspirin usage. The variable Aspirin indicates the daily aspirin regimen while Statin denotes a subject' statin medication. Additionally, \textsc{psa} represents the total antigen level circulating in a subject’s blood, measured in ng/mL. For patients sensitive to Statin medications, the aim is to determine how to manipulate relevant risk factors to minimize both Statin and \textsc{psa}. To this end, we treat both Statin and \textsc{psa} as target variables. The treatment variables include \textsc{bmi}, Weight, \textsc{ci}, and Aspirin usage with interventional domains $\mathcal{D}(\textsc{bmi}) = [20,30]$, $\mathcal{D}(\text{Weight})=[50,100]$, $\mathcal{D}(\textsc{ci})=[-100,100]$ and $\mathcal{D}(\text{Aspirin})= [0,1]$. We choose to consider a specific age groups of interest, and define $U_{\text{age}}$ as a Gaussian random variable with mean 65 and standard deviation 1, focusing on individuals close to the age of 65. The single-objective version of \textsc{Health}, aiming to minimize only \textsc{psa}, has previously been used to demonstrate the applicability of \textsc{cbo} \citep{CBO}, as well as for several of its variants (e.g. \citet{pmlr-v216-gultchin23a} and \citet{pmlr-v202-aglietti23a}). 

\end{document}